\documentclass[11pt]{article}

\usepackage[preprint]{acl}

\usepackage{times}
\usepackage{latexsym}
\usepackage{float}
\newfloat{listing}{t}{lol}
\floatname{listing}{Listing}

\usepackage[T1]{fontenc}

\usepackage[utf8]{inputenc}
\usepackage{microtype}

\usepackage{inconsolata}

\usepackage{graphicx}
\usepackage{amsmath}
\usepackage{arydshln}
\definecolor{hpink}{HTML}{F7E1ED}
\usepackage{latexsym}
\usepackage{float}
\usepackage[utf8]{inputenc} 
\usepackage[T1]{fontenc}    
\usepackage{url}            
\usepackage{booktabs}       
\usepackage{wrapfig}
\usepackage{amsfonts}       
\usepackage{nicefrac}       
\usepackage{microtype}      
\usepackage{xcolor}         
\usepackage{multirow}
\usepackage{graphicx}
\usepackage{makecell} 
\usepackage{booktabs}
\usepackage{multirow}
\usepackage{color, colortbl}
\usepackage{subcaption}
\usepackage{tabu}
\usepackage{pifont}
\usepackage{makecell}
\usepackage{threeparttable}
\usepackage{enumitem}
\usepackage{hhline}
\usepackage{scalerel,xparse}
\usepackage{ulem}
\usepackage{caption}
\usepackage{pgfplots}
\usepackage{diagbox}
\usepackage{array}
\usepackage{color, colortbl}
\definecolor{Gray}{gray}{0.5}
\definecolor{LGray}{gray}{0.9}
\definecolor{darkblue}{RGB}{94,110,186}
\definecolor{darkGreen}{RGB}{92, 148, 110}
\definecolor{myblue}{RGB}{14, 121, 178}
\definecolor{myred}{RGB}{192, 0, 0}

\usepackage{adjustbox}
\usepackage{amssymb}
\usepackage{lipsum}  
\usepackage{pdfpages}

\usepackage{booktabs}
\usepackage{multirow}
\usepackage{xcolor,colortbl}
\definecolor{rowhl}{RGB}{232,238,250}
\definecolor{grouprow}{RGB}{245,245,247}
\usepackage{subcaption}
\definecolor{rowhl}{RGB}{232,238,250}
\definecolor{grouprow}{RGB}{245,245,247}
\definecolor{deltacol}{RGB}{15,110,86}

\usepackage[table]{xcolor}   
\definecolor{headerbg}{RGB}{235,235,238}
\usepackage{booktabs}
\usepackage{subcaption}
\usepackage{xcolor,colortbl}
\definecolor{rowhl}{RGB}{232,238,250}

\usepackage{listings}
\usepackage{xcolor}

\definecolor{codegreen}{rgb}{0.0,0.5,0.1}   
\definecolor{codegray}{rgb}{0.5,0.5,0.5}
\definecolor{codepurple}{rgb}{0.58,0.0,0.55} 
\definecolor{codeblue}{rgb}{0.13,0.29,0.53}  
\definecolor{backcolour}{rgb}{0.97,0.97,0.95}

\lstdefinestyle{mypython}{
  language=Python,
  backgroundcolor=\color{backcolour},
  commentstyle=\color{codegreen}\itshape,
  keywordstyle=\color{codeblue}\bfseries,
  stringstyle=\color{codepurple},
  basicstyle=\ttfamily\small,
  numberstyle=\tiny\color{codegray},
  breaklines=true,
  captionpos=b,
  showstringspaces=false,
  numbers=left,
  numbersep=8pt,
  frame=single,
  rulecolor=\color{codegray},
  literate={→}{$\rightarrow$}{1}
}


\usepackage{fontawesome5}

\usepackage{pifont}

\definecolor{mydarkgreen}{rgb}{0.2,0.7,0.2}

\definecolor{darkergreen}{RGB}{0,100,0}
\definecolor{myred}{rgb}{1,0,0}
\definecolor{myblue}{rgb}{0,0,1}
\definecolor{mygreen}{rgb}{0,1,0}
\definecolor{myyellow}{rgb}{1,1,0}

\definecolor{my_green}{RGB}{51,102,0}
\definecolor{my_red}{RGB}{204, 0, 0}
\renewcommand{\checkmark}{\textcolor{my_green}{\ding{51}}} 

\title{HPP: Hierarchical Programmatic Probing for Long Video Understanding by Decoupling Perception and Reasoning }

\author{Awais Rauf \\
  Queen's University Belfast\\
United Kingdom\\
  \texttt{a.rauf@qub.ac.uk} \\\And
  Ahmed Hasssan\\
  AMD\\
  United States \\
  \texttt{Ahmed.hasssan@amd.com} \\\And
  Greg Slabaugh \\
  Queen Mary University of London \\
  United Kingdom \\
  \texttt{g.slabaugh@qmul.ac.uk}  \\
\AND
\faGlobe\ \texttt{\href{https://awaisrauf.com/HPP}{AwaisRauf.com/HPP}}
  }

\begin{document}
\maketitle
\begin{abstract}
Understanding long videos requires fine-grained perception and multi-step, higher-order reasoning over complex, long-range spatio-temporal dynamics. Vision-language models (VLMs) encode video frames into visual tokens and attempt to perform both perception and multi-step planning latently, within a single forward pass. This coupled formulation, however, is bottlenecked by the LLM's limited capacity to discover and execute multi-step strategies in its latent representations. To address this bottleneck, we propose Hierarchical Programmatic Probing (HPP), a framework that decouples semantic perception from higher-order temporal reasoning by reformulating long video understanding as iterative, programmatic exploration of a hierarchically segmented video. Specifically, a coding-capable LLM plans and executes a multi-step strategy in an interactive coding environment, probing the video for information and invoking a VLM for localized perception on demand. To make probing tractable over long videos, we introduce three components: information-density-aware hierarchical segmentation, late-interaction semantic retrieval, and structured probing functions for coarse-to-fine temporal localization. We validate HPP on LongVideoBench, which requires both fine-grained perception and long-range relational reasoning, and show that decoupling the two via iterative programmatic probing yields substantial gains. Further results on EgoSchema, VideoMME, and MLVU demonstrate the effectiveness of our approach across diverse long-video benchmarks. Relevant code and supporting materials are available at: \href{https:awaisrauf.com/hpp}{https:awaisrauf.com/hpp}. 
\end{abstract}

\section{Introduction}

Long video understanding has become increasingly critical as video content accounts for a growing share of global internet traffic across diverse domains.
The dominant paradigm extends Large Language Models (LLMs) to video by projecting frames into the LLM's embedding space via a visual encoder and performing perception and reasoning jointly over the resulting token sequence within a single forward pass~\cite{minigpt4-2023, wang2023all, llava2023,lin2023video0llava0,bai2025qwen2050vl,wang2025internvl3050}.

This monolithic formulation of VLMs is structurally mismatched for long video understanding.
Unlike short clips, long videos require both localized perceptual grounding to extract sparse visual information from thousands of redundant frames and multi-step planning over long temporal horizons to reason about complex spatio-temporal interactions. However, recent work has shown that latent reasoning within a single forward pass is ineffective for resolving temporal dependencies and executing multi-step planning~\cite{li2025temporal, xu2026depth}. Compounding this, the number of visual input tokens grows rapidly with duration (e.g., one hour video produces 1.08M visual tokens in Gemini~\cite{gemini_video}), inflating compute and KV-cache as LLMs scale while diluting relevant signal across the extended context~\cite{liu2024lost}.

Video-agent frameworks aim to address these challenges by treating videos as text-like documents, relying on dense captioning of sampled frames~\cite{wang2023lifelongmemory, zhang2024simple, kahatapitiya2025language} or incorporating rigid search processes and external tools into predefined workflows~\cite{videoagent2024, wang2025videotree, yang2025vca}. For example, \cite{kahatapitiya2025language} converts video chunks into a language repository through captioning and summarization, enabling downstream reasoning with LLMs, while VideoTree~\cite{wang2025videotree} does dense captioning of a tree structured video frames, inputting to LLM for reasoning in a single step. This reliance on dense captioning and manually designed heuristics imposes rigid workflows that constrain adaptive exploration, limiting the ability to fully exploit both the sparse nature of video data and the iterative, dynamic problem-solving capabilities of modern LLMs. Meanwhile, these capabilities of LLMs have advanced rapidly recently~\cite{carlos2024swebench, openai2026gpt53codex, qwen3_next_coder, anthropic2026claudeopus}. 
LLMs have become significantly better at coding and structured problem-solving, exhibiting strong planning, multi-step execution, and emergent symbolic reasoning~\cite{anthropic2025subagents, schroeder2025thread, yang2025emergent, zhang2025recursive}. These abilities enable them to autonomously decompose tasks and execute iterative strategies through code, making them particularly well-suited for long-horizon, programmatic reasoning.

\begin{figure*}[t]
    \centering
    \includegraphics[width=1\linewidth]{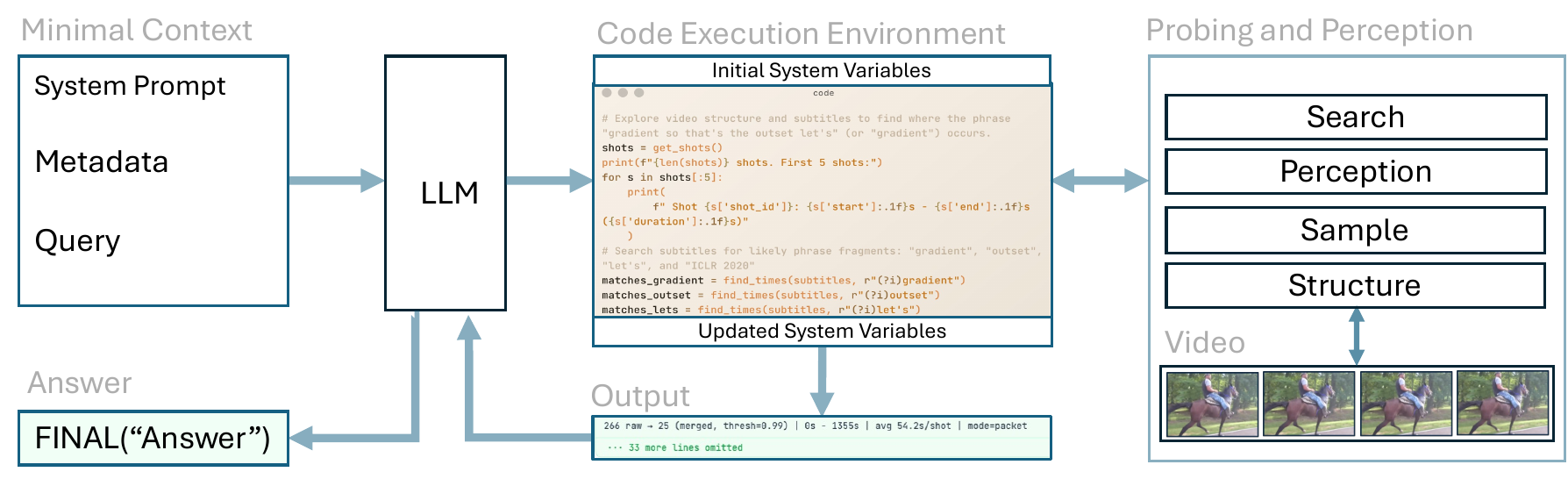}
    \caption{
\textbf{Overview of HPP.} A coding-capable LLM is initialized with minimal context and iteratively navigates a video through code execution and structured probing functions in a REPL, invoking a VLM on demand for targeted perception before producing the final answer.}
    \label{fig:main}
\end{figure*}

Building on these programmatic and multi-step planning capabilities, we ask: can long-video understanding be reformulated by decoupling semantic perception from higher-order temporal reasoning, allowing a coding-capable LLM to plan and execute structured exploration while a VLMs handles event recognition and fine-grained perceptual extraction within frames or short segments? 
To this end, we cast long video understanding as iterative, programmatic probing over structured video segmentation. 
We propose Hierarchical Programmatic Probing for long video understanding (HPP), a framework that enables a coding-capable LLM to perform high-level reasoning while interactively exploring long videos within a coding environment. Within this framework, the LLM is responsible for strategy formulation, planning, and relational inference through programmatic interaction, whereas a lightweight VLM is invoked for targeted semantic perception over selected frames or short segments.

Video navigation through code, however, presents unique challenges compared to text: videos consist of massive, highly redundant frame sequences with varying information density, making naive frame-level processing computationally prohibitive.
To enable efficient exploration, our framework provides the LLM with a set of structured probing functions inside the programming environment.
Specifically, we introduce (1) an information-density-aware, computationally efficient hierarchical structuring method that converts raw frames into multi-scale segments by exploiting temporal redundancy; (2) a motion-adaptive late-interaction semantic retrieval module for localizing query-relevant segments; (3) structure-aware video sampling functions for fine-grained temporal navigation; and (4) a small VLM for targeted spatio-temporal understanding of segments.

To demonstrate the effectiveness of HPP, we first evaluate our approach on LongVideoBench~\cite{wu2024longvideobench}, a large-scale long-video benchmark that stratifies videos by duration and task complexity across perception and relational reasoning. Consistent with our central hypothesis, decoupling semantic perception from programmatic reasoning enables strong long-horizon performance even with a smaller VLM. 
We further validate our approach on general long video understanding benchmarks, including EgoSchema, VideoMME, MLVU~\cite{egoschema2023,videomme2024,zhou2024mlvu}, confirming the effectiveness of our proposed HPP across diverse long-video settings.

\section{Related Work}
\label{sec:related_work}
Here, we briefly position our work in relation to prior research; a more detailed discussion is provided in Appendix~\ref{supp_sec:related_work}.

\noindent\textbf{Long Video Understanding. }
Prior post-training works for long video understanding span a wide range of methods, including extending context to accommodate more visual tokens~\cite{zhang2024longva} and compressing visual tokens~\cite{shen2024longvu,videollama,liu2025video,tao2025dycoke}, often trading fine-grained details for efficiency. More relevant to our work, agent-based frameworks address long-context limitations through tool-augmented reasoning or predefined search procedures~\cite{chen2025lvagent, wang2025videotree, yang2025vca}, but typically rely on dense captioning, rigid heuristics, or multiple coordinated VLMs.  
In contrast, our method decouples perception from reasoning, reformulating long-video understanding as hierarchical programmatic probing in which an LLM navigates the video through code and adaptively discovers strategies for complex, long-horizon tasks.

\noindent\textbf{Coding-based Approaches for Vision Tasks. }
Neural Module Networks (NMNs)~\cite{andreas2016neural,hu2017learning,johnson2017inferring,hu2018explainable} pioneered compositional visual question answering, decomposing each question into a program layout that is then executed over the image. However, these approaches require learning program layouts. More recent methods have used LLM's code generation abilities to solve image understanding tasks in single forward pass\cite{suris2023vipergpt, gupta2023visual, marsili2025visual}. {In contrast, our approach is focused on long video understanding by decoupling perception and reasoning, enabling LLMs to perform multi-step planning with code.}

\section{Method}
Our method consist of an LLM navigating video through code execution and a set of efficient video probing functions. 
To fully exploit the LLM’s reasoning and coding abilities without imposing a rigid task formulation, we initialize the LLM with minimal inputs: a system prompt for context, video metadata, and the user query. The model is given access to (i) an interactive Python environment (REPL) initalized with system variables, (ii) a set of video probing functions for structured navigation, and (iii) a small vision-language model (VLM) for targeted spatio-temporal understanding.

Given a query, the reasoning-and-navigation LLM first interprets the task and formulates a high-level strategy. It then executes this strategy by writing and executing Python code, selectively invoking probing functions to navigate the video and querying the VLM when fine-grained visual understanding is required. This iterative loop of reasoning, code execution, and video interaction enables explicit symbolic reasoning while efficiently exploring long videos, as illustrated in Figure~\ref{fig:main}.

\begin{figure*}[t]
    \centering
    \includegraphics[width=0.53\linewidth]{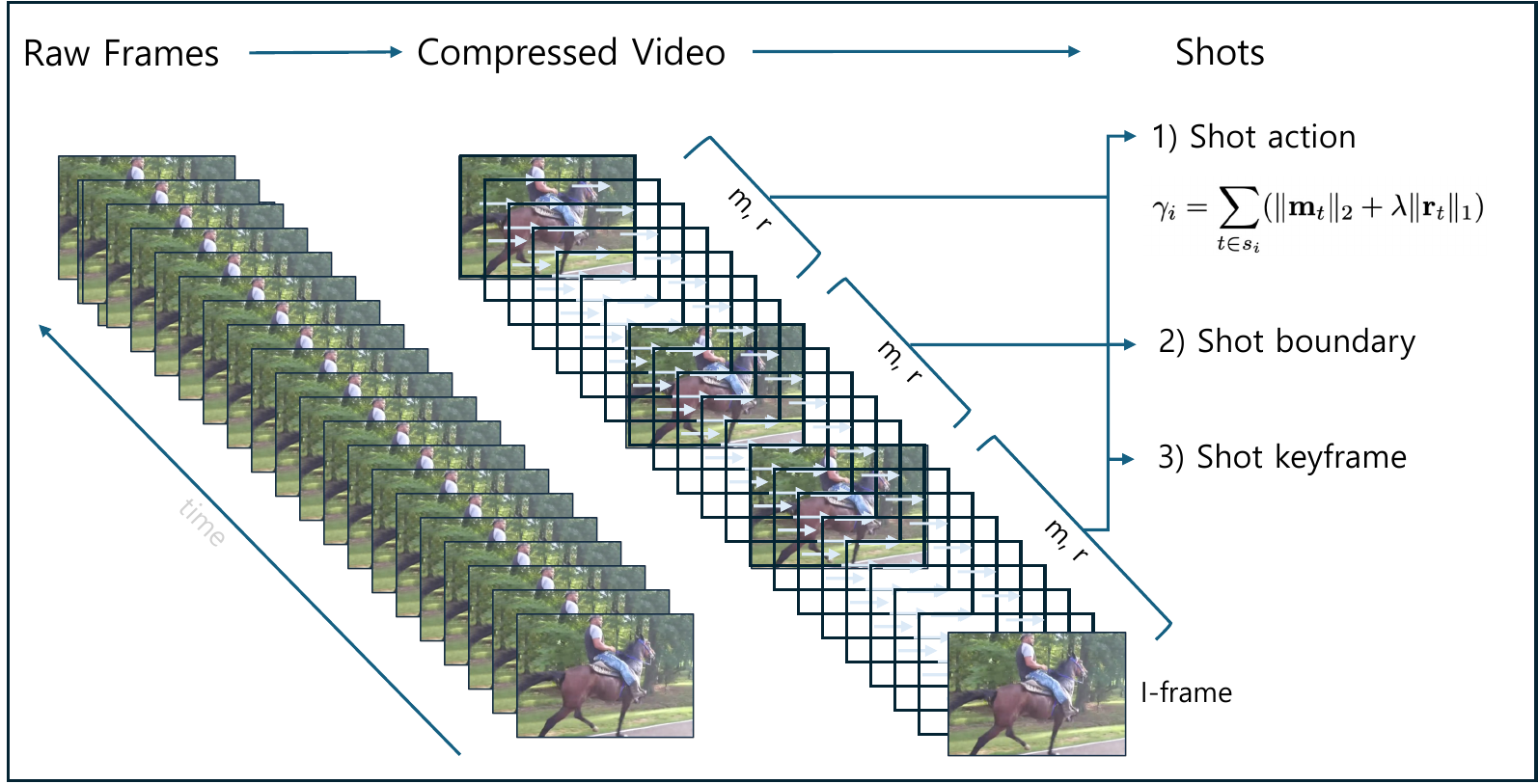}
    \hfill
    \includegraphics[width=0.42\linewidth]{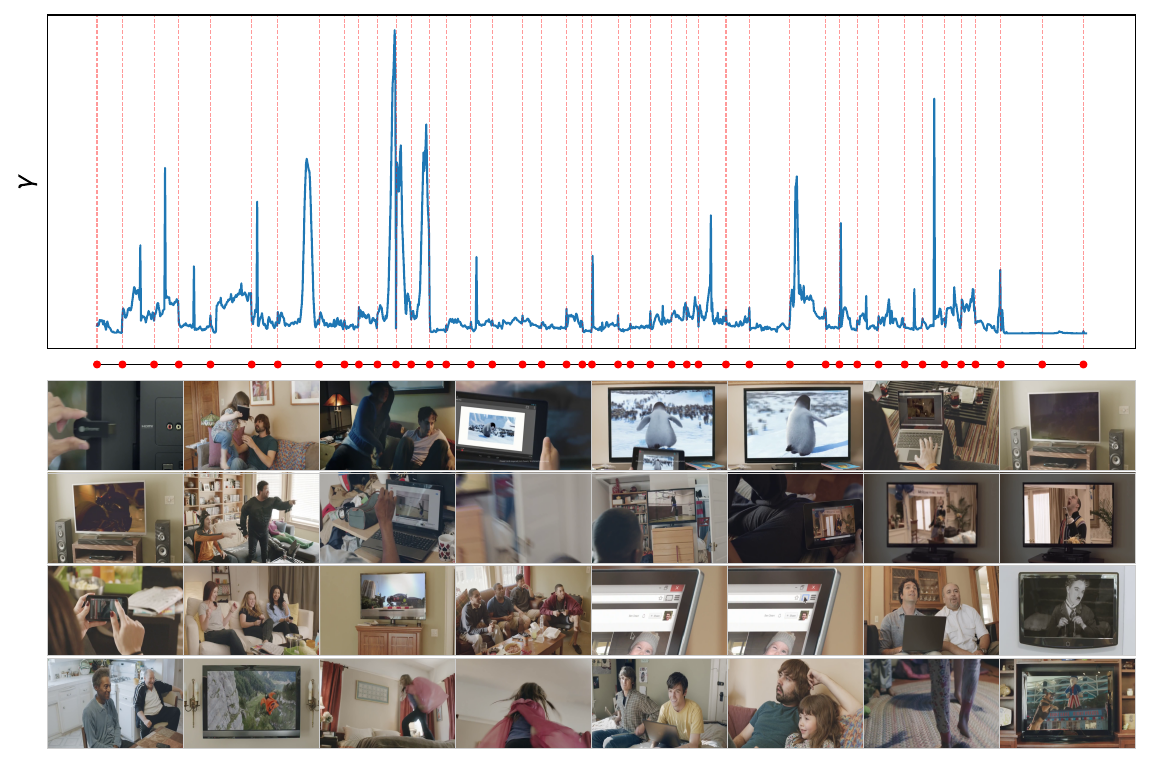}
    \caption{(Left) Overview of our method for efficiently converting raw video frames into shot-level segments by directly leveraging codec information. (Right) An example of information-density-aware segmentation of a video into shots with keyframes.}
    \label{fig:shots}
\end{figure*}

\subsection*{Interactive Hierarchical Programmatic Probing Environment}
The core of our framework is an interactive probing environment that enables a coding-capable LLM to iteratively navigate long videos, retrieve query-relevant evidence, and refine its reasoning based on intermediate execution feedback. Inspired by Recursive Language Models~\cite{zhang2025recursive}, we adopt a REPL-based interaction paradigm in which the LLM executes code within an isolated container and receives structured, terminal-style outputs, forming a tight reasoning–execution loop.

Unlike natural language, videos consist of dense and highly redundant frame sequences, making naive frame-level exploration computationally prohibitive. To support efficient and structured probing, we design four key modules within the REPL environment. First, we introduce a computationally efficient hierarchical structuring mechanism that leverages codec information to transform raw videos into multi-scale segments. Second, we design a semantic retrieval module that allows the LLM to localize query-relevant segments using both global embeddings and fine-grained token–frame late-interaction similarity. Third, we expose structured video sampling functions that enable fine-grained temporal navigation while constraining exploration to a controlled action space. Finally, we integrate a lightweight VLM that can be queried for targeted spatio-temporal understanding over selected frames or clips, while higher-level reasoning remains within the LLM.

\noindent \textbf{1) Hierarchical Structure for Efficient Probing.}

Most VLMs implicitly assume that videos consist of static information and downsample frames at uniform intervals. However, this uniform sampling is suboptimal as videos exhibit significant temporal redundancy and substantial variation in information density: some segments contain critical events, while others are largely repetitive. To enable the LLM to navigate video more efficiently in the interactive coding environment, we propose a hierarchical module that decomposes long videos into progressively coarser, perceptually and temporally coherent segments as needed.

The first part is to efficiently partition large number of video frames ($\{f_t\}_{t=1}^{T}$) into shots ($\{s_i\}_{i=1}^{N}$) with similar content and small motion. 
The temporal motion can be extracted by handcrafted~\cite{beauchemin1995computation} or deep learning-based optical flow~\cite{teed2020raft}. However the expensive computation of optical flow methods makes them unsuitable for our purpose. 
To perform this step efficiently, we leverage compressed representation used for storage and transmission and can be extracted at high speed~\cite{wu2018compressed,jin2024video}. 
Specifically, most modern video codecs (e.g., MPEG-4, H.264 etc.) exploit temporal redundancy of successive frames by only storing few frames that are least compressible (I-frames) and reconstruct rest of the frames from motion vectors and residuals~\cite{musmann2005advances}. 
From an information-theoretic perspective, compression aims to minimize the expected code length of a video sequence by reducing the conditional entropy,
$ \sum_{t=1}^{T} H(f_t \mid \hat{f}_{t}),
$
where $\hat{f}_{t}$ is motion-compensated prediction of frame $f_t$ and stores motion vector $\mathbf{m}_t$ and residuals $\mathbf{r}_t$, such that
$
\hat{f}_t \approx \mathcal{W}(f_{t-1}, \mathbf{m}_t, \mathbf{r}_t)
$
where $\mathcal{W}(\cdot)$ denotes motion warping.

\begin{figure*}[t]
    \centering
    \includegraphics[width=0.9\linewidth]{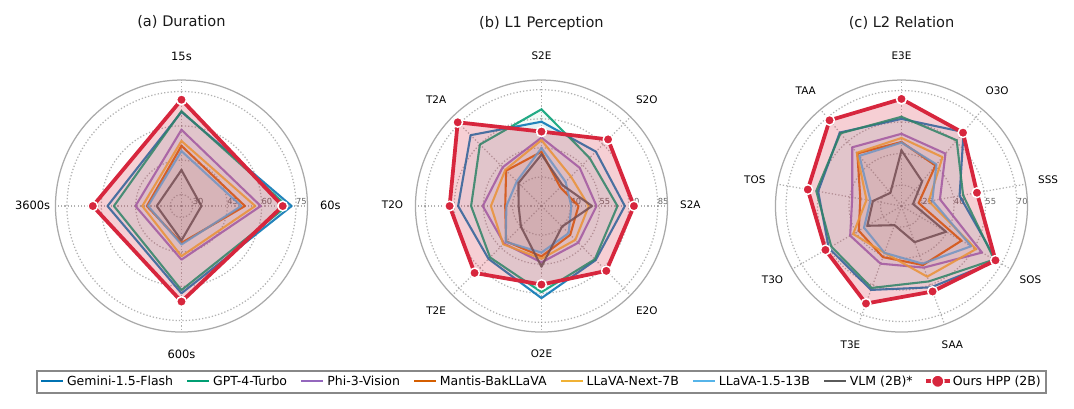}
    \caption{
     \textbf{HPP vs. VLMs on LongVideoBench}, broken down by video duration, perception-based questions (L1), and long horizon relational questions (L2). Results are taken from ~\cite{wu2024longvideobench}, $^\star$ indicate evaluation under the setting in which VLM is used in our method. }
    \label{fig:longvideobench}
\end{figure*}

As shown in Figure~\ref{fig:shots}, we leverage information from the compressed bitstream as a coarse but extremely efficient structural prior. We use timestamps of I-frames ($\{t_k\}_{k=1}^{N}$) to partition the frames into shot candidates
$
s_i = \{f_t \mid t_k \leq t < t_{k+1}\}, 
$
where $f_{t_k}$ is keyframe for $k$-th shot. 
To further reduce redundancy, we merge adjacent shots whose keyframes are visually similar ($\text{merge}_{s_k, s_{k+1}} \text{if } \|f_{t_k} - f_{t_{k+1}}\| < \tau_{s}$).
For each shot $s_k$, we compute a shot energy using motion and residual vectors,
\begin{equation}
    \gamma_i = \sum_{t \in s_i} (\|\mathbf{m}_t\|_1 + \lambda \|\mathbf{r}_t\|_1),
    \label{eq:gamma}
\end{equation}
which serves as a proxy for information change, as video codecs allocate fewer bits to more redundant content while preserving essential temporal context.~\cite{lu2019dvc} This proxy is used in subsequent steps for motion-adaptive frame selection.

For longer videos, shot-level segmentation alone is insufficient for effective navigation. To address this, shots can be further grouped into spatio-temporally coherent scenes and globally semantically consistent clusters using a lightweight distilled visual encoder. We defer the details to Section~\ref{supp_sec:method} of the Appendix.

\noindent \textbf{2) Fine-grained , Motion-aware Semantic Search for Query-based Localization. }
After dividing a video into manageable segments, the next step is to enable the LLM to retrieve query-relevant content via text. Video retrieval methods typically employ a bi-encoder architecture, in which frames are uniformly sampled from a video segment, encoded using a vision encoder, and aggregated into a single global embedding~\cite{shu2024videoxl, xu2021videoclip}. This global representation, together with a text embedding, is then used to compute similarity and retrieve relevant results.

However, unlike general video retrieval, text-based localization within a single video requires a more fine-grained approach, as different segments may share substantial visual similarity. For example, localizing a segment corresponding to the query “bird flying over water” in a nature documentary requires distinguishing it from visually similar scenes of birds flying over a desert. Moreover, motion dynamics can vary significantly across short segments, making uniform frame sampling suboptimal. To address these challenges, we design a video localization method that adaptively samples frames based on the magnitude of motion within each shot and computes similarity using late interaction between independently encoded shot and query embeddings~\cite{khattab2020colbert, reddy2025video}.

The localization is performed at the segment level (e.g., shots). Let $\{s_i\}_{i=1}^{N}$ denote the set of shots for a video, where each shot $s_i$ contains $T_{s_i}$ frames $\{f_{i, t}\}_{t=1}^{T_{s_i}}$ and a keyframe. We sample frames proportionally to the shot's motion energy $\gamma_i$ computed using Eq.~\ref{eq:gamma}. The number of sampled frames for the $i$-th shot is given by
$\eta_i = \frac{\gamma_i}{\sum_{m=1}^{N} \gamma_m} \, B_{\text{total}},$
where $B_{\text{total}}$ is total frame budget.  
For each shot $s_i$, $\eta_i$ frames $\{f_{i,k}\}_{k=1}^{\eta_i}$ are sampled and encoded using a CLIP-like vision encoder $\phi_v$ to obtain frame-level embeddings
$
\mathbf{z}_{i, t} = \phi_v(f_{i,t}) \in \mathbb{R}^{d}.
$
Similarly, a textual query consisting of $n_t$ tokens ${q_l}_{l=1}^{n_t}$ is encoded using a text encoder $\phi_t$ to obtain token-level embeddings
$
\mathbf{q_l} = \phi_t(q_l) \in \mathbb{R}^{d}.
$

To capture fine-grained token–frame interactions within each shot while maintaining the efficiency of dot-product similarity, we compute a weighted late-interaction MaxSim score:
\[
\text{MaxSim}(q, s_j) 
= 
\frac{1}{\eta_i} 
\sum_{k=1}^{\eta_i} 
w_{j,k} \cdot 
\max_{i} 
\left( 
\mathbf{q}_i^\top \mathbf{z}_{j,k} 
\right),
\]
where $w_{j,k}$ is the weight of the $k$-th frame in shot $s_i$, assigning higher importance to keyframes. This per-token motion-adaptive per-frame interaction captures fine-grained and dynamic spatio-temporal information.  
To further capture shot-level global information, we aggregate frame and query embeddings via mean pooling:
$
\mathbf{z}_i 
= 
\frac{1}{\eta_i} 
\sum_{t=1}^{\eta_i} 
\mathbf{z}_{j,t},
$
and $
\mathbf{q} 
= 
\frac{1}{n_t} 
\sum_{i=1}^{n_t} 
\mathbf{q}_i,
$
and compute a shot-level similarity with the query:
$
\text{Sim}(q, s_j) 
= 
\mathbf{q}^\top \mathbf{z}_i.
$
The final similarity score combines fine-grained and global similarities:
$
\text{score}(q, s_j) 
= \alpha_s \cdot \text{MaxSim}(q, s_j)+ (1-\alpha_s) \cdot \text{Sim}(q, s_j),
$
where $\alpha_s$ is a weighting hyperparameter. 

By modeling both global shot-level semantics and fine-grained token–frame interactions, the search module captures static spatial context as well as motion-driven action cues, leading to more precise and reliable localization.
Top-$k$ candidate shots are returned along with their similarity scores, which LLM can use to decide which shots to further probe.  
Scenes are not embedded separately; instead, a scene representation is computed by using the embeddings of its constituent shots, reducing computation while preserving semantic structure.

\noindent \textbf{ 3) Effective Sampling Interface and Perception.} 
After localizing relevant temporal regions, the next step is to enable the LLM to efficiently sample relevant segments for effective perception with the VLM. We provide a set of structured functions that support multi-scale temporal sampling. These functions include retrieving a frame at a specific timestamp, extracting frames over a time interval, hierarchy-aware frame sampling, temporal zoom-in within a selected segment, and segment clipping. More details of these functions are provided in Section~\ref{supp_sec:probing_functions}. These structured functions provide a predictable and well-defined interface within the REPL, enabling the LLM to systematically explore the video at the segment level with greater reliability and control.

\begin{table}[t]
    \centering
    \caption{\textbf{Decoupling perception from reasoning along two axes on LongVideoBench.} }
    \label{tab:hpp_axes}
    \vspace{-5pt}
    \begin{subtable}{\columnwidth}
        \centering
        \caption{\textbf{Better Perception Yields Better Performance. }
        With the fixed LLM, HPP consistently improves over monolithic VLMs at both small and large vision-encoder scales, demonstrating the ability of programmatic reasoning to utilize better perception models. }
        \vspace{-5pt}
        \label{tab:vlm_axis}
        \setlength{\tabcolsep}{5pt}
        \renewcommand{\arraystretch}{1.22}
        \resizebox{\columnwidth}{!}{%
            \begin{tabular}{l l cc cc c c}
            \toprule
            \multirow{2}{*}{\textbf{Setting}} 
                & \multirow{2}{*}{\textbf{Vision}}
                & \multicolumn{2}{c}{\textbf{LVB-3600}}
                & \multicolumn{2}{c}{\textbf{Avg}}
                & \multirow{2}{*}{\textbf{All}}
                & \multirow{2}{*}{$\boldsymbol{\Delta}$} \\
            \cmidrule(lr){3-4} \cmidrule(lr){5-6}
                & & L1 & L2 & L1 & L2 & & \\
            \midrule
            \rowcolor{grouprow}\multicolumn{8}{c}{\textit{\textbf{Small vision Encoder} ($\sim$0.3B)}} \\
            Monolithic VLM    & 2B  & --   & --   & --   & --   & 57.4 & --                      \\
            \rowcolor{rowhl}
            + HPP (Kimi-K2)   & 2B  & 70.5 & 56.0 & 70.5 & 70.1 & \textbf{70.3} & \textcolor{deltacol}{$+10.9$} \\
            \addlinespace[2pt]
            \rowcolor{grouprow}\multicolumn{8}{c}{\textit{\textbf{Large Vision Encoder} ($\sim$6B)}} \\
            Monolithic VLM    & 38B    & --   & --   & --   & --   & 65.7 & --                      \\
            \rowcolor{rowhl}
            + HPP (Kimi-K2.6)   & 38B    & 69.5 & 65.7 & 76.5 & 69.2 & {\color{red}\textbf{73.5}} & \textcolor{deltacol}{$+7.8$}  \\
            \bottomrule
            \end{tabular}}
    \end{subtable}

    \vspace{0.5em}

    \begin{subtable}{\columnwidth}
        \centering
        \caption{\textbf{Coding ability of the LLM transfers to video reasoning.}
        With the same VLM (2B VLM, 0.3B encoder), stronger coding LLMs yield directly stronger video understanding, demonstrating the effectiveness of programmatic reasoning for long video understanding.}
        \label{tab:llm_axis}
        \setlength{\tabcolsep}{5pt}
        \renewcommand{\arraystretch}{1.22}
        \resizebox{\columnwidth}{!}{%
            \begin{tabular}{l c cc cc c}
            \toprule
            \multirow{2}{*}{\textbf{LLM}}
                & Term
                & \multicolumn{2}{c}{\textbf{LVB-3600}}
                & \multicolumn{2}{c}{\textbf{Avg.}}
                & \multirow{2}{*}{\textbf{All}} \\
            \cmidrule(lr){3-4} \cmidrule(lr){5-6}
                & World & L1 & L2 & L1 & L2 & \\
            \midrule
            Baseline VLM & - & - & - & -  & -& 57.4$^\star$\\
            \midrule
            MiniMax-M2.5   & 49.0& 60.0 & 42.7 & 63.7 & 54.7 & 59.2 \\
            \rowcolor{rowhl}
            Kimi-K2.6        & {57.5} & {68.0} & {60.0} & {69.1} & {67.6} & {68.3} \\
            \rowcolor{rowhl}
            Claude Opus 4.6 & 62.5 & 76.0 & 72.0 & 80.3 & 72.3 & {\color{red}76.3}\\
            \bottomrule
            \end{tabular}}
    \end{subtable}
\end{table}

\section{Effectiveness of Decoupled Perception and Reasoning}
\label{sec:results_sec4}
To evaluate effectiveness of our central thesis, that decoupling perception from reasoning and enabling multi-step code-based planning improves long-video understanding, we first compare HPP against VLMs on LongVideoBench (LVB)~\cite{wu2024longvideobench}, which stratifies questions into perceptual (L1) and relational (L2) categories. While L1 questions primarily test perceptual grounding such as recognizing objects, actions, and events within local segments, \textit{L2 demands relational reasoning over temporally distributed evidence, requiring both fine-grained perception and long-horizon multi-step reasoning across long videos}. LVB consist of 3600 questions spanning four duration groups from short to longer. 

\begin{figure}[t]
    \centering
\includegraphics[width=1\columnwidth,height=0.85\textheight,keepaspectratio]{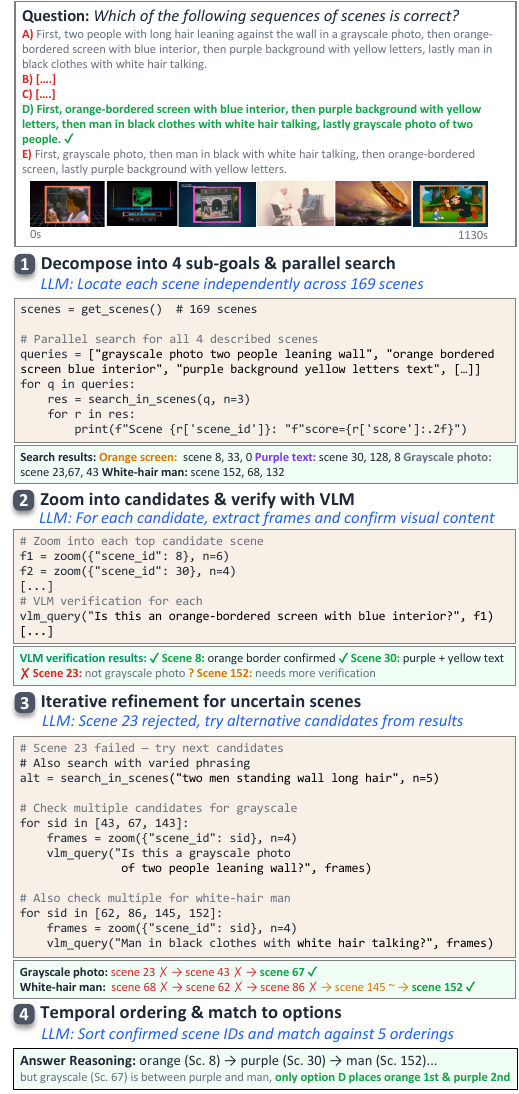}
    \caption{\textbf{An example of HPP in action.} The model decomposes the query into sub-goals and executes its strategy in the REPL using compositional code (e.g., parallel search across paraphrased queries). HPP probes the video efficiently despite its 34k frames: structuring it into 169 scenes, retrieving and zooming into localized regions for targeted inspection, and iteratively refining its evidence before answering.}
    \label{fig:example2}
    \vspace{-15pt}
\end{figure}

Figure~\ref{fig:longvideobench} reports results on LVB, with baseline results taken from~\cite{wu2024longvideobench}. For cost efficiency, we use GPT-5-mini as the reasoning LLM and Qwen3-VL-2B~\cite{bai2025qwen30vl} as the perception VLM. On shorter videos and L1 task, our method performs comparably to VLMs. 
However, our method's gains are most pronounced on longer videos (600s and 3600s subset) and L2 relational tasks that require multi-step planning and multimodal temporal information retrieval. For example, TAA (attribute change of an object between two subtitle-anchored moments) and T3O (identifying an object appearing before/after a subtitle) both demand cross-modal temporal anchoring followed by fine-grained perception at distant segments, precisely the kind of reasoning that monolithic VLMs struggle with, as they must integrate information across thousands of frames in a single pass. Similarly, strong improvements on E3E (temporal ordering of events) and SOS (re-identifying an object across different scenes) demonstrate that our hierarchical localization, programmatic probing and multi-step planning enable the model to first localize relevant moments and then invoke targeted perception, reducing distraction from redundant frames and enabling structured long-range understanding.

\paragraph{Decoupling enables independent scaling of VLM and LLM.}
To further test our decoupling hypothesis, we scale either the vision encoder or the reasoning LLM (Table~\ref{tab:hpp_axes}) on a 600-question subset of LongVideoBench, sampled proportionally across the original distribution of 4 duration groups, 2 difficulty levels, and 13 question types.
\textit{Varying the vision encoder} and keeping LLM fixed (Kimi-K2.6~\cite{team2026kimi}), HPP improves the monolithic baseline by $+10.9$ at 0.3B vision encoder and $+7.8$ at 6B vision encoder.We report vision-encoder size since VLMs often scale the LLM without correspondingly scaling the encoder. For this comparison, we use InternVL3.5-2B and InternVL3.5-38B~\cite{wang2025internvl3050}. 
%
%
Furthermore, to show the benefit of stronger coding models, we switch LLM with increasing coding abilities (as measured by TerminalWorld\cite{chu2026terminalworld}). We use three models (Kimi-K2.6, MiniMax-m2.7~\cite{minimax2026m27} and Claude Opus4.6~\cite{anthropic2026claudeopus46}) and fixed VLM (Qwen3VL-2B). 
Results are reported in Table~\ref{tab:hpp_axes}(b), with performance scaling monotonically with general coding ability ($59.2 \rightarrow 68.3 \rightarrow 76.3$), confirming that HPP directly benefits from stronger LLMs.

\begin{listing}
\vspace{-5pt}
\begin{lstlisting}[style=mypython, numbers=none, caption={\textbf{An example of effective utility of programmatic probing}. HPP employes 5--15 paraphrase variants of the target concept and uses the REPL to merge and de-duplicate results by timestamp, yielding a more complete retrieval set than any singlequery.}, label={code:pattern2}]
# Question:  In this video, how many instances are there of the 'stomping grapes' action scene in total?
# Stage 2 - Localization
queries = ["grape stomping", "stomping", "grapes", "feet in grapes", ...,"wine press", "winery harvest"]
all_hits, seen = [], set()
for q in queries:
    for h in search_in_shots(q, n=30):
        key = (round(h['start'], 2), round(h['end'], 2))
        if key not in seen:
            seen.add(key)
            all_hits.append(h)
print(f"Total unique candidate shots: {len(all_hits)}")
\end{lstlisting}
\end{listing}

\noindent \textbf{How Code-based Programmatic Probing Works.} Figure~\ref{fig:example2} illustrates HPP on a 4-scene temporal ordering question from a 20-minute history documentary, where reused visual templates (e.g., colored borders) create many partial matches across 169 scenes. The LLM decomposes the question into four sub-goals, issues parallel semantic searches, zooms into candidates for VLM verification, and iteratively refines failed matches with varied query phrasings and cross-modal cues. After confirming all four scenes, it reasons over their temporal order to select the correct answer, inspecting only 68 of 33,909 frames (0.2\%).

\noindent \textbf{Emergent Coding-based Strategies.} A notable property of HPP is that the reasoning LLM spontaneously develops effective coding strategies to efficiently navigate the video, without being explicitly instructed to do so. For instance, Listing~\ref{code:pattern2} shows a pattern we term \texttt{multi-query ensemble search with deduplication}: since semantic search is nearly free after the first invocation (embeddings are cached), the LLM exploits this by issuing multiple paraphrase variants of the target concept (synonyms, compound descriptions, and visual surface forms) at negligible additional cost. It then merges and deduplicate results by timestamp, forming a retrieval set more complete than any single query alone. Such patterns arise consistently across different questions and video types because of flexibility offered by code-based navigation. We provide a detailed analysis and additional examples in Section~\ref{supp_sec:emergent}.

\section{General Long Video Understanding }
 \begin{table}[t]
\centering
\caption{Comparison on challenging long-video understanding benchmarks. Our method outperforms or achieves competitive results compared with methods relying on significantly larger vision models.}
\label{tab:results}
\renewcommand{\arraystretch}{1.05}
\scalebox{0.6}{
\begin{tabular}{l cccc}
\toprule
\textbf{Model}
& \textbf{Ego}
& \textbf{LVB}
& \textbf{MLVU}
& \textbf{VMME} \\
\midrule
Claude 4.5 Opus           & --   & 57.3 & --   & 77.6 \\
GPT-4o~\cite{gpt4o}       & 72.2 & 66.7 & 64.6 & 77.2 \\
Gemini 1.5 Pro~\cite{gemini15} & 71.1 & 64.0 & --   & 81.3 \\
\midrule
ShareGPT4Video-8B~\cite{chen2024sharegpt4video} & --   & 39.7 & 46.4 & 43.6 \\
VideoChat2-7B~\cite{mvbench}                    & 56.7 & 39.3 & 47.9 & 43.8 \\
LongVA-7B~\cite{longva}                         & --   & 51.3 & 58.8 & 54.3 \\
LLaVA-Video-7B~\cite{llavavideo}                & 57.3 & 58.2 & 70.8 & 69.7 \\
\midrule
Aria-28B~\cite{aria}                            & --   & 64.2 & 72.3 & 72.1 \\
LLaVA-Video-72B~\cite{llavavideo}               & 65.6 & 64.9 & --   & 76.9 \\
InternVL-2.5-78B~\cite{internvl2.5}             & --   & 63.6 & 75.7 & 74.0 \\
LLaVA-OneVision-72B~\cite{li2025llavaonevision} & 62.0 & 63.2 & 68.0 & 69.6 \\
VideoLLaMA-2-72B~\cite{videollama2}             & 63.9 & --   & 45.6 & 64.7 \\
\midrule
VideoAgent~\cite{videoagent2024}                & 54.1 & --   & --   & --   \\
VideoTree~\cite{wang2025videotree}              & 61.1 & 41.3 & --   & --   \\
Dr.\ Video~\cite{ma2025drvideo}                 & 61.9 & --   & --   & 71.7 \\
VideoRAG-72B~\cite{luo2026video}                & --   & 65.4 & 73.8 & 75.7 \\
\midrule
\rowcolor{gray!20}
{Ours HPP}                               & {70.4} & {66.4} & 69.3 & 71.8 \\
\bottomrule
\end{tabular}
}
\end{table} 
\label{lab:SOTA}
To further demonstrate the effectiveness of HPP for long-video understanding, we compare it against a broad spectrum of approaches across four challenging benchmarks (Table~\ref{tab:results}). 
The baselines include large proprietary models(e.g., GPT-4o, Gemini 1.5 Pro), open-source VLMs ranging from 7B–78B parameters, and recent agentic frameworks such as VideoAgent~\cite{wang2025videoagent} and VideoRAG~\cite{videorag}. For HPP, we pair GPT-5-mini as the coding-capable LLM with Qwen3VL-2B as the lightweight perception model. Our method achieves strong performance across all benchmarks. For instance, on LongVideoBench, our framework achieves 66.4 average accuracy, outperforming several larger models. Comparable improvements on EgoSchema and MLVU, where HPP matches or surpasses models an order of magnitude larger, confirm that the benefit generalizes across long-video settings. These results support our central hypothesis: externalizing higher-order temporal reasoning into adaptive, code-driven probing rather than scaling perception within a single forward pass, yields strong long-video understanding without scaling the VLM.

\begin{figure}
    \centering
    \includegraphics[width=1\linewidth]{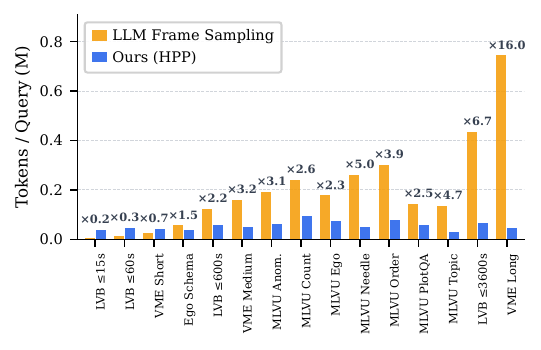}
    \caption{Comparison of average token usage with standard LLM frame sampling~\cite{gemini_video} and HPP. }
    \label{fig:resource_tokens}
\end{figure}

\noindent \textbf{Comparison of Duration and Resource Usage. }
Figure~\ref{fig:resource_tokens} compares HPP's token usage against standard frame sampling (e.g., Gemini~\cite{gemini_video}) across all 15 sub-tasks of our four evaluation datasets, ordered along the x-axis by increasing video duration and task complexity. While HPP uses slightly more tokens on short videos (15–60s) due to its probing overhead, the trend reverses sharply as videos grow longer and tasks more complex: reaching up to 16$\times$ fewer tokens than standard sampling on the longest, hardest tasks.
Appendix~\ref{supp_sec:resources} provides a detailed analysis of HPP's resource usage across video lengths and question types, covering API costs, LLM/VLM token consumption, probing-function patterns, and REPL utilization. The analysis shows that HPP adaptively allocates compute to query difficulty: few iterations on easy queries, longer probing on temporally complex ones. It also scales substantially more efficiently than standard frame sampling.

\noindent \textbf{System Component Ablation. }
We conduct our ablation study on a 120 sample subset from the 3600-second split of LongVideoBench, as it represents the most challenging setting, requiring both long-video understanding and long-horizon reasoning. We first analyze the contribution of the major components of our framework (bottom of Table~\ref{tab:ablations}). 
The standalone VLM used in our framework performs poorly on its own, as expected, since HPP uses it with 8 frames per clip. 
Similarly, disabling either the our semantic localization module or the structured probing functions significantly degrades accuracy (46.2 and 48.9 respectively), showing that both components are necessary for effective temporal localization and organization of retrieved evidence. Removing the VLM entirely leads to the largest degradation (29.2). We also observe that prompt design and model capacity affect performance: using a simplified prompt (full prompt is in Appendix Section~\ref{supp_sec:sys_prompt}) reduces accuracy (59.2 vs. 63.7), while scaling the VLM from 2B to 8B improves performance (66.2). using a stronger LLM (GPT5) further increases accuracy to 68.3.

%

\begin{table}[t]
\centering
\setlength{\tabcolsep}{5pt}
\renewcommand{\arraystretch}{1.1}
\newsavebox{\bottomtbl}
\sbox{\bottomtbl}{%
\begin{tabular}{llcccr}
\hline
\rowcolor{headerbg}
\textbf{LLM} & \textbf{Prompt} & \textbf{Search} & \textbf{Struct.} & \textbf{VLM} & \textbf{LVB} \\
\hline
$\times$ & $\times$ & $\times$ & $\times$ & 2B & 30.5$^\star$ \\
\hline
GPT-mini & Default & \checkmark & \checkmark & 2B & 63.7 \\
GPT-mini & Default & $\times$   & \checkmark & 2B & 46.2 \\
GPT-mini & Default & \checkmark & $\times$   & 2B & 48.9 \\
GPT-mini & Default & \checkmark & \checkmark & $\times$ & 29.2 \\
GPT-mini & Default & $\times$   & $\times$   & 2B & 48.9 \\
GPT-mini & Simple  & \checkmark & \checkmark & 2B & 59.2 \\
\hline
GPT-mini & Default & \checkmark & \checkmark & 8B & 66.2 \\
GPT      & Default & \checkmark & \checkmark & 8B & \textbf{68.3} \\
\hline
\end{tabular}}
\resizebox{0.95\columnwidth}{!}{%
\makebox[\wd\bottomtbl][s]{%
\begin{tabular}{lc}
\hline
\rowcolor{headerbg}
\textbf{Embedding Model} & \textbf{Acc.} \\
\hline
None    & 46.2 \\
SigLIP  & 47.5 \\
CLIP    & 58.3 \\
VideoXL & 63.7 \\
\hline
\end{tabular}%
\hfill
\begin{tabular}{lc}
\hline
\rowcolor{headerbg}
\textbf{Similarity Metric} & \textbf{Acc.} \\
\hline
Sim                     & 57.9 \\
MaxSim                  & 54.8 \\
Hybrid ($\alpha_s$,0.3) & 53.9 \\
Hybrid ($\alpha_s$,0.7) & 63.7 \\
\hline
\end{tabular}}}
\vspace{4pt}
\resizebox{0.95\columnwidth}{!}{\usebox{\bottomtbl}}
\vspace{-10pt}
\caption{Retrieval and system-component ablations. $^\star$ indicate VLM evaluated on settings used in HPP.} 
\label{tab:ablations}
\vspace{-15pt}
\end{table}
\noindent \textbf{Semantic Search Ablation. }
The upper side of Table~\ref{tab:ablations} shows our evaluation for design choices in the semantic search, including the embedding model and similarity metric for retrieval. The embedding model plays a crucial role in identifying relevant video segments. Without embeddings, performance drops to 46.2, while stronger multimodal representations progressively improve results, with VideoXL~\cite{shu2024videoxl} achieving the best performance (63.7). We also compare similarity metric for retrieval: cosine similarity achieves an accuracy of 57.9, token-level MaxSim performs worse (54.8), while our proposed hybrid approach achieves 63.7, highlighting effectiveness of design choices.

\section{Conclusion}

We presented HPP, a programmatic framework for long-form video understanding that explicitly decouples semantic perception from higher-order temporal reasoning. Instead of treating videos as having uniformly distributed information, our approach reformulates video understanding as iterative code-driven probing over hierarchically structured video representations within a coding environment. In this setting, a coding-capable LLM performs planning, strategy execution, and relational reasoning, while a lightweight VLM is invoked only for targeted perceptual understanding. This design enables efficient exploration of long videos and better leverages the emerging reasoning capabilities of modern LLMs.

\section{Limitations}

While HPP is effective across a range of long-video benchmarks, several limitations remain. 
Firstly, HPP's effectiveness is fundamentally tied to the coding LLM that drives probing. Our scaling experiments (Table~\ref{tab:hpp_axes}) show that performance scales monotonically with the LLM's coding ability: HPP inherits future gains from stronger LLMs but is also bounded by the LLM's current capacity to decompose queries, formulate adaptive strategies, and recover from failed probes. 

Although externalizing reasoning into code circumvents the depth limit of latent multi-step reasoning, each individual probing decision still relies on the LLM's intrinsic planning ability within a single turn. Weaker LLMs may issue ill-formed retrieval queries, terminate prematurely without sufficient evidence, or fail to chain probes coherently. More subtly, HPP also inherits the LLM's biases: query interpretation, the choice of paraphrases for semantic retrieval, and the prompts issued to the VLM all reflect the LLM's pretraining priors, which can systematically privilege some types of evidence or framings over others. 

Another limitation arises from HPP's autonomous, multi-turn nature. Because the LLM plans and executes its own probing strategy, trajectories occasionally go awry in ways that are difficult to anticipate from the question alone. For instance, the model may latch onto an incorrect hypothesis early and confirm it through biased follow-up probes, repeatedly issue near-duplicate retrieval queries, or commit to a final answer before sufficient evidence has been gathered. Another failure mode observed in with some smaller LLMs is model's inability to return final answer and instead keep looping for more and more confirmations.

A related failure mode is context filling: across multiple turns, intermediate reasoning, VLM responses and especially verbose code outputs accumulate in the context window, and we occasionally observe the LLM weighting earlier (sometimes weakly supported) evidence over more recent and more reliable findings, or losing track of the original query under accumulated context. These failures are characteristic of autonomous agentic systems and are not fully mitigated by the structured probing API. 
Finally, our evaluation focuses on multiple-choice long-video QA benchmarks (LongVideoBench, EgoSchema, VideoMME, MLVU). Performance on open-ended generation, dense captioning, fine-grained temporal grounding, or interactive video tasks remains to be characterized.

\section{Acknowledgments}
This work was supported by the Engineering and Physical Sciences Research Council [grant number EP/Y009800/1], through funding from Responsible Ai UK (KP0016). 

\bibliography{main}
\clearpage
\appendix
\let\oldsection\section
\tableofcontents

\section{Extended Method Details}
\label{supp_sec:method}

In this section, we briefly expand on three key aspects of the REPL environment: hierarchical segmentation, VLM-based perception, and the various probing functions exposed by the environment.

\begin{figure*}[t]
    \centering
    \includegraphics[width=1\linewidth]{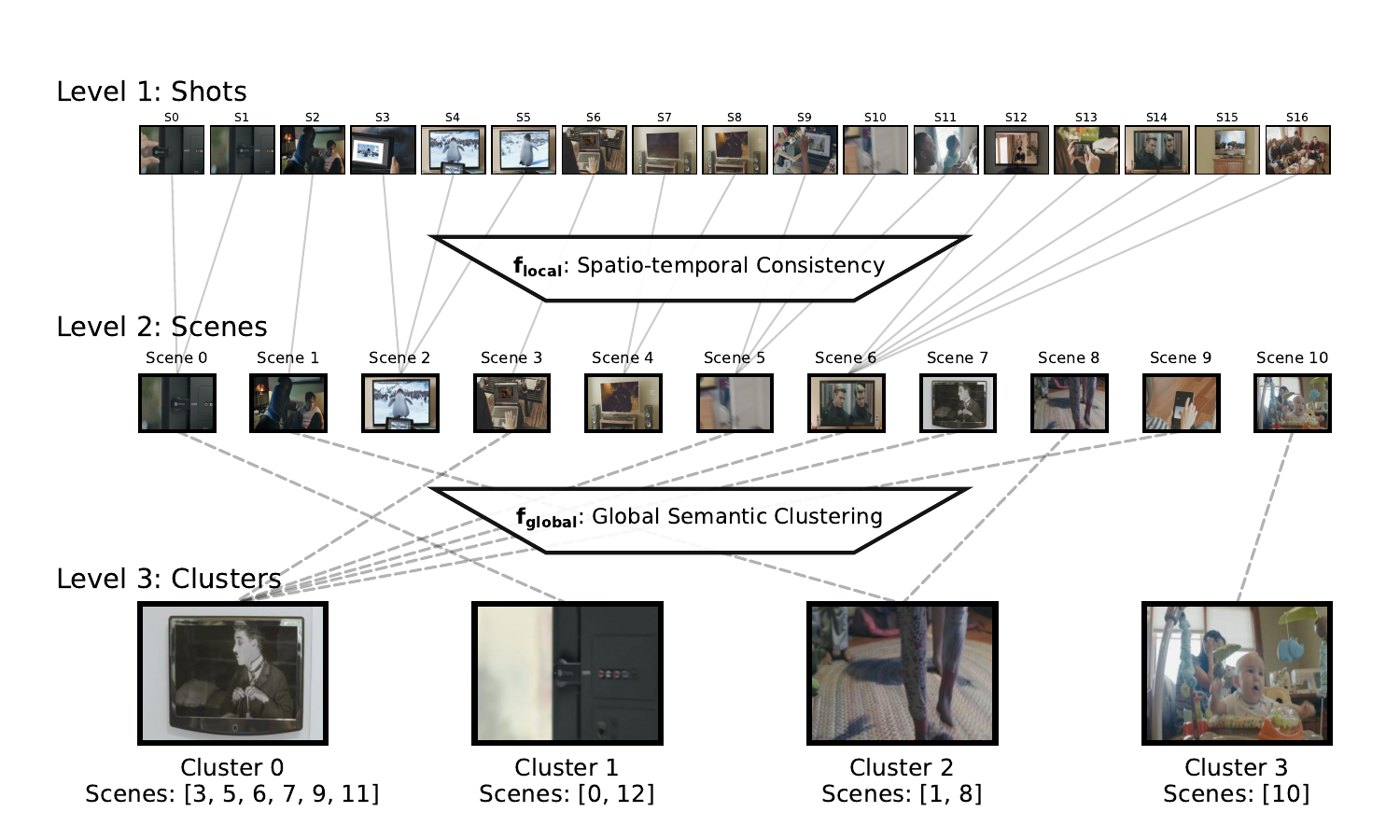}
    \caption{Overview of our method for constructing hierarchical video units from shots to scenes, and clusters using a lightweight image encoder.  
    }
    \label{fig:scenes_clusters}
\end{figure*}

\label{supp_sec:shot_scenes_vlm}
\noindent\textbf{From Shots to Scenes and Clusters.}
Shot-level decomposition enables the LLM to efficiently inspect a video at a fine-grained level. However, videos often contain semantically similar scenes, including temporally distant segments that convey the same underlying narratives.
To this end, hierarchical structure module can group shots into temporally consistent scenes ($\{c_l\}_{l=1}^{L}$) and semantically similar clusters ($\{\mathcal{C}_{r}\}_{r=1}^{R}$), as shown in Figure~\ref{fig:scenes_clusters}.

For scenes, each shot $s_i$ with keyframe $\{f_{t_{k}}\}$ is encoded using a small distilled visual encoder $\Phi(\cdot)$\cite{fanzinger2024CVPR} to obtain spatio-temporal and semantic features
$
\mathbf{z}_{t_k}, \mathbf{e}_{t_k} = \Phi(f_{t_k}) \in \mathbb{R}^{D}.
$
Given the sequence of shot-level features $\{\mathbf{z}_i\}_{i=1}^{N}$, we compute pairwise cosine distances between adjacent shots:
\[
\mathcal{D}_{s}(i, i+1) = 1 - \frac{\mathbf{z}_i^\top \mathbf{z}_{i+1}}{\|\mathbf{z}_i\| \|\mathbf{z}_{i+1}\|},
\]
and contiguous shots with high similarity ($\mathcal{D}(i, i+1) \leq \tau_s$) are grouped into scenes.

Next, non-contiguous and semantically similar scenes are grouped into clusters. Given global semantic embeddings for a shot calculated in previous step, we pool to form scene level embeddings 
$\mathbf{h}_l =\dfrac{1}{c_l} \sum_{i \in c_l } \mathbf{e}_{i} $ and compute a pairwise similarity matrix using cosine similarity:
$
\mathbf{S}(i, j) = 1 - \frac{\mathbf{h}_i^\top \mathbf{h}_j}{\|\mathbf{h}_i\|_2 \|\mathbf{h}_j\|_2}.
$
The resulting $L \times L$ similarity matrix is used as input to agglomerative clustering to group semantically related scenes into clusters $\{\mathcal{C}_r\}_{r=1}^R$. Each cluster is represented by the union of keyframes from its constituent scenes and forms the highest level of the hierarchy, enabling coarse-to-fine navigation of long videos.

\noindent \textbf{Segment-level Perception with Small VLM.}
To perform localized spatio–temporal understanding over the extracted segments, we employ a lightweight VLM (e.g., Qwen3VL-2B~\cite{qwenvl-2023}) as a perception module. The LLM generates task-conditioned textual prompts and submits them, together with the sampled frame(s) or short video clip, to the VLM in order to extract query-relevant visual and temporal evidence.
The VLM is invoked without additional system prompts or external contextual information; it functions purely as a visual–temporal encoder and interpreter. Higher-level reasoning, cross-segment aggregation, and decision-making remain entirely within the LLM.

\noindent \textbf{REPL Environment and Structured Probing Functions}
\label{supp_sec:probing_functions}
HPP exposes the video to the coding LLM through a stateful Python Read--Evaluate--Print Loop (REPL), in which the LLM emits code, observes printed output, and continues reasoning across turns. Each session is initialized with a small set of pre-loaded variables that anchor the LLM in the task without consuming context budget: a \texttt{subtitles} list of \{start, end, text\} dictionaries (or \texttt{None} when unavailable), and video metadata exposed via \texttt{get\_video\_info()} (duration, resolution, frame rate, shot count). These initial variables let the LLM scope the problem before issuing any query and act as a free, always-available textual information about the video. Crucially, the REPL is {persistent across turns}: the LLM can accumulate evidence in arbitrary Python variables (e.g., \texttt{evidence\_by\_option = \{\}}), revisit and refine retrieved segments, and compose primitives into multi-step strategies that span many iterations. This persistence is what makes adaptive, code-driven probing tractable. Specifically, intermediate findings do not need to be re-derived, and the LLM's reasoning state is externalized into inspectable variables rather than carried latently inside a single forward pass.

Furthermore, HPP exposes the video to the coding LLM as system functions in REPL, organized into five families that span the cost: information trade-off and together support coarse-to-fine temporal localization.

\begin{figure*}[t]
    \centering
    \includegraphics[width=0.85\linewidth]{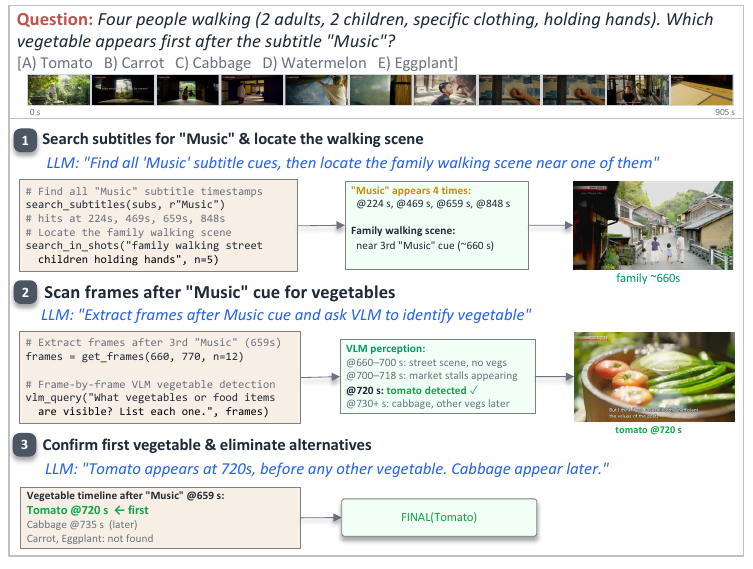}
    \caption{\textbf{An example of HPP in action.} The model decomposes the query into sub-goals, executes its strategy in the REPL, and probes the video. It first searches for the word "music" and adjacent family-walking scenes, then uses the VLM to identify the target. Despite the video containing 27,000 frames, the model efficiently navigates it by first structuring the video into shots, retrieves relevant evidence, and invokes the VLM for targeted perception only on selected scenes. }
    \label{fig:placeholder}
\end{figure*}

\begin{itemize}
    \item \textbf{Video-structure functions} (\texttt{get\_shots}, \texttt{get\_scenes}, \texttt{get\_clusters}, \texttt{get\_shot\_detail}) return the hierarchical segmentation as text---shot and scene boundaries, keyframe timestamps, and visually similar clusters ---letting the LLM orient itself temporally before committing any perceptual budget. 
    
    \item \textbf{Semantic retrieval functions}  (\texttt{search\_in\_shots}, \texttt{search\_in\_scenes}) perform late-interaction queries to localize segments relevant to a natural-language phrase, returning ranked candidates with timestamps and scores. 
    
    \item \textbf{Subtitle functions} (\texttt{search\_subtitles}, \texttt{find\_times}, \texttt{get\_subtitles}, \texttt{summarize\_subtitles}) provide free, text-only access to dialogue and on-screen factual content, which often answers dialogue or factual questions outright. 
    
    \item \textbf{Frame and clip extraction} (\texttt{get\_frame\_at}, \texttt{get\_frames}, \texttt{sample}, \texttt{get\_overview}, \texttt{zoom}, \texttt{sample\_shot}, \texttt{get\_clip}) materialize visual content from chosen regions at controlled granularity, with \texttt{zoom} and \texttt{sample} operating directly over the hierarchical segments returned by the structure functions. 
    
    \item \textbf{Vision-language queries} (\texttt{vlm\_query}) route a small set of frames or a short clip to a lightweight VLM with a natural-language prompt, returning a textual description---the only path through which raw pixels enter the LLM's reasoning.
\end{itemize}

\renewcommand{\arraystretch}{1.2}
\begin{table*}[t]
\setlength{\tabcolsep}{6pt}

\caption{Results on the MLVU benchmark. Our method significantly improves performance across holistic, single-detail, and multi-detail long-video understanding tasks compared to open and close source single pass models. }
\label{tab:overall:test}
\centering
\scalebox{0.80}{
\begin{tabular}{lcccccccc}
\toprule
\specialrule{0em}{0.3pt}{0.3pt}
\multirow{2}{*}{\textbf{Methods}}  &
\multicolumn{2}{c}{\textbf{Holistic}} &
\multicolumn{3}{c}{\textbf{Single Detail}} &
\multicolumn{2}{c}{\textbf{Multi Detail}} &
\multirow{2}{*}{\textbf{M-Avg}} \\
\specialrule{0em}{0.3pt}{0.3pt}
\cmidrule(r){2-3} \cmidrule(r){4-6} \cmidrule(r){7-8}
\specialrule{0em}{0.1pt}{0.1pt}
& TR & AR & NQA & ER & PQA & AO & AC & ~ \\
\specialrule{0em}{0.3pt}{0.3pt}
\hline

MovieChat~\cite{moviechat2023} & 18.7 & 10.3 & 23.3 & 15.1 & 16.0 & 17.1 & 15.0 & 16.5 \\
Movie-LLM~\cite{moviellm2024} & 27.5 & 25.6 & 10.0 & 11.3 & 16.0 & 20.0 & 21.7 & 18.9 \\
LLaMA-VID~\cite{li2024llama} & 20.9 & 23.1 & 21.7 & 11.3 & 16.0 & 18.6 & 15.0 & 18.1 \\
MA-LMM~\cite{malmm2024} & 44.0 & 23.1 & 13.3 & 30.2 & 14.0 & 18.6 & 13.3 & 22.4 \\
MiniGPT4-Video~\cite{minigpt4video-2024} & 64.9 & 46.2 & 20.0 & 30.2 & 30.0 & 15.7 & 15.0 & 31.7 \\
LongVA~\cite{zhang2024longva} & 81.3 & 41.0 & 46.7 & 39.6 & 46.0 & 17.1 & 23.3 & 42.1 \\
Video-CCAM~\cite{fei2024videoccam} & 79.1 & 38.5 & 45.0 & 52.8 & 56.0 & 24.3 & 26.7 & 46.1 \\
LLaVA-Onevision~\cite{li2024llavaov} & 83.5 & 56.4 & 46.7 & 58.4 & 58.0 & 35.7 & 23.3 & 51.7 \\
GPT-4o$^\dag$~\cite{gpt4o} & 83.7 & 68.8 & 42.9 & 47.8 & 57.1 & 46.2 & 35.0 & 54.5 \\

VLM Baseline (2B) & 81.4 & 63.5 & 63.1 & 50.3 & 56.2 & 36.7 & 21.8 & 54.5 \\

\rowcolor{gray!20}
Ours HPP (2B) & 79.8 & 64.5 & 70.4 & 60.8 & 70.9 & 81.1 & 53.9 & 69.3 \\

\bottomrule
\end{tabular}
}
\vspace{-10pt}
\end{table*}

The functions are organized along the cost : coding and printing are free, subtitle operations and structure queries are cheap, frame extraction and semantic retrieval are moderate, and VLM calls, particularly over clips, are the most expensive. This structure pushes the LLM toward a natural coarse-to-fine strategy: orient using cheap signals (structure, subtitles, retrieval), narrow the temporal region of interest, and invoke perception only on the resulting targeted segments. Because the functions are exposed as Python functions within the REPL, the LLM composes them adaptively, chaining retrieval into zoom into VLM, maintaining evidence in variables across turns, and probing multiple regions in parallel, rather than following a predefined workflow. The same primitives thus support a wide range of query types, from dialogue-driven factual questions answered entirely through subtitles to fine-grained perceptual questions resolved by clip-level VLM queries.

\section{Extended Results}

\subsection{Extended Results on Benchmark Datasets}
Table~\ref{tab:overall:test} reports results on the MLVU benchmark, which evaluates long-video understanding across three categories: holistic reasoning (Topic Reasoning or TR, Anomaly Recognition or AR), single-detail queries (Neddle QA or NQA, Ego Reasoning or ER, Plot Question or PQA), and multi-detail reasoning (Action Order or AO, Action Count or AC). Our approach achieves a 69.3 overall average, substantially outperforming the baseline VLM (54.5) and surpassing strong open-source and proprietary models such as LLaVA-OneVision (51.7) and GPT-4o (54.5). The improvements are particularly pronounced in multi-detail reasoning, where our method achieves 81.1 on Action Order and 53.9 on Action Count, far exceeding prior methods that typically remain below 50. These tasks require aggregating multiple pieces of information across distant segments, which standard end-to-end VLMs struggle with due to temporal redundancy and limited reasoning structure. By contrast, HPP’s hierarchical probing allows the model to iteratively localize relevant segments and combine evidence across time. We also observe strong gains on single-detail tasks, including Needle QA (70.4) and Plot QA (70.9), demonstrating improved temporal grounding and question-specific retrieval.

\begin{figure*}[t]
    \centering
    \includegraphics[width=1\linewidth]{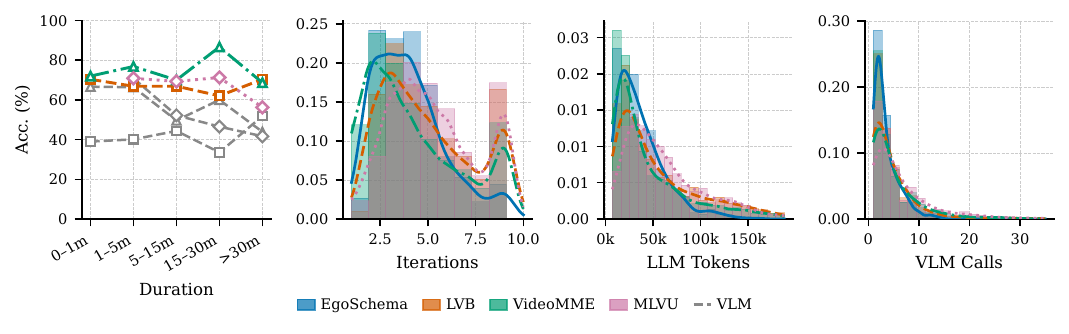}
    \caption{Performance comparison with the baseline VLM across video duration groups, along with the distribution of resource usage (iterations, LLM tokens, and VLM calls). }
    \label{fig:accuracy_vs_resource}
\end{figure*}

\subsection{Visualization of Reasoning Traces}

Our framework answers queries by iteratively interacting with the video through a programmatic interface. During inference, the LLM plans actions, writes and executes code in the REPL, invokes probing functions, and combines the returned information with its internal reasoning to progressively refine the answer. To make this process transparent and easier to inspect, we provide an interactive visualization application as part of supplementary code.

The information on how to run the application attached in Readme file in the code. Running the application launches a dashboard that summarizes key statistics about the model's execution traces across datasets. Users can select a dataset and split to explore individual reasoning sessions in detail. Each session can be viewed from two complementary perspectives: a functional view that highlights the sequence of probing function calls and retrieved information, and a coding view that displays the generated code and its execution results. These visualizations allow readers to follow how the model explores the video, narrows down candidate segments, and gathers evidence before producing the final answer.

\section{Cost Comparison and Resource Utilization}
\label{supp_sec:resources}
In this section, we present a detailed analysis of the costs associated with our method compared to using GPT-5-mini Vision. We also provide a comprehensive breakdown of resource utilization. To structure this analysis, we divide it into four parts: (1) A comprehensive comparison of performance and video duration, distribution of number of coding iterations, distribution of LLM tokens and distribution of VLM token usage across four datasets, (2) a comparison of API costs of our method and using GPT-5-mini's native vision understanding, (3) a breakdown of token usage of our method for each task, (4) a comparison of the average utilization of the different probing functions within HPP, and (5) an evaluation of REPL environment usage.

\subsection{Comparison of Duration and Resource Usage }
In Figure~\ref{fig:accuracy_vs_resource}, we compare the accuracy of our method with the baseline across different video duration groups and analyze the corresponding resource utilization. First, our method maintains consistent performance as video duration increases, demonstrating robust long-horizon understanding. Second, the computational resources used by the framework adapt to the complexity of the query and the length of the video. For example, on EgoSchema, most queries are resolved within 2–3 iterations with relatively few VLM calls, whereas on LVBench, where questions often require more complex temporal reasoning, the framework performs longer iterative probing. This behavior highlights the adaptive nature of our approach, which allocates computation only when necessary.

\begin{figure*}[t]
    \centering
    \includegraphics[width=1\textwidth]{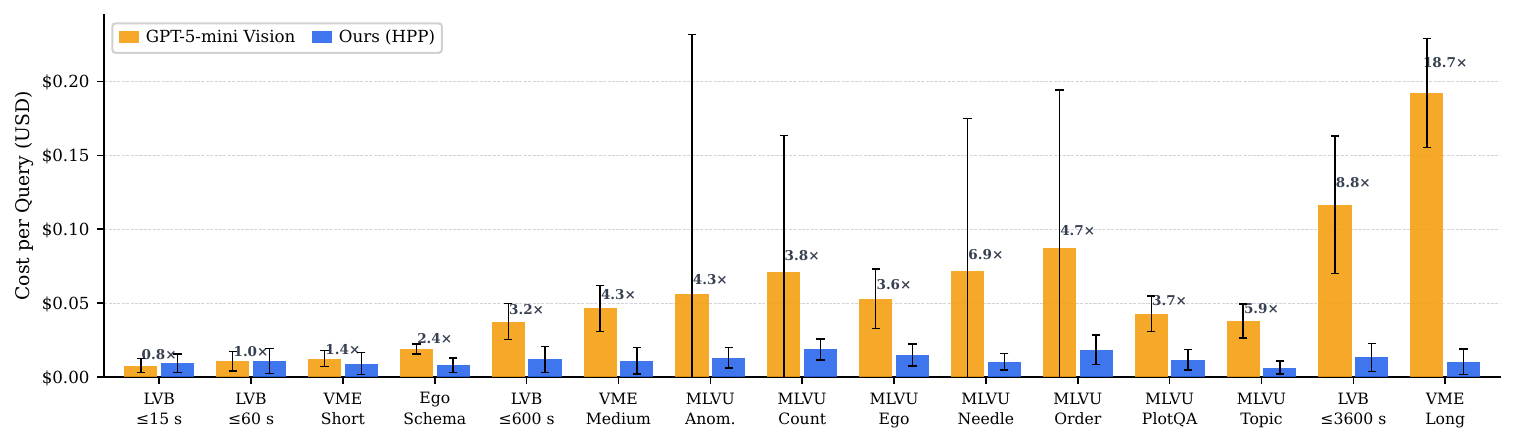}
  \caption{Comparison of API usage cost between our method and GPT-mini's native vision. 
Costs are reported for different split of four benchmark datasets used in our method and visualized. 
While both methods have comparable cost for short videos, our approach scales more efficiently and results in substantially lower API cost for longer videos.}
    \label{fig:api_cost}
\end{figure*}
\subsection{Comparison of API Usage Cost}

Figure~\ref{fig:api_cost} compares the API usage cost of our method with GPT-mini's native vision capabilities in order to evaluate the efficiency of our approach. For clarity, the results are presented using both linear and logarithmic scales. As shown in Figure~\ref{fig:api_cost}, the cost of our method remains comparable to the baseline for short videos, while becoming significantly lower as video length increases. This demonstrates that our approach scales more efficiently for long-form video understanding. 

During our experiments, we log both input and output token usage for each query. To estimate the overall API cost of our method, we assume that approximately $70\%$ of the input tokens correspond to the system prompt, which can be cached across requests. Cached tokens incur a significantly lower cost compared to regular input tokens. Following the pricing model reported by OpenAI~\cite{openai_api_pricing}, we compute the total cost using the following formula:

\[
\begin{aligned}
\text{Cost}
&= \text{input\_tokens}_{M} \cdot (0.8\,C_{\text{cached}} + 0.2\,C_{\text{input}}) \\
&\quad + \text{output\_tokens}_{M} \cdot C_{\text{output}}
\end{aligned}
\]

where $\text{input\_tokens}_{M}$ and $\text{output\_tokens}_{M}$ denote the number of input and output tokens in millions, and $C_{\text{cached}}$, $C_{\text{input}}$, and $C_{\text{output}}$ correspond to the respective token prices.

Recent LLM-based video understanding systems typically process video by sampling frames at 1 FPS, where each frame requires approximately 300 tokens~\cite{gemini_video}. For a one-hour video, this results in $300 \times 3600 = 1.08$M tokens for visual inputs alone. We compare this baseline with the number of tokens used by our method across different video scales, demonstrating the improved efficiency of our approach. 

\subsection{Token Usage Across Video Lengths and Question Types}

\begin{figure*}[t]
    \centering
    \includegraphics[width=1\textwidth]{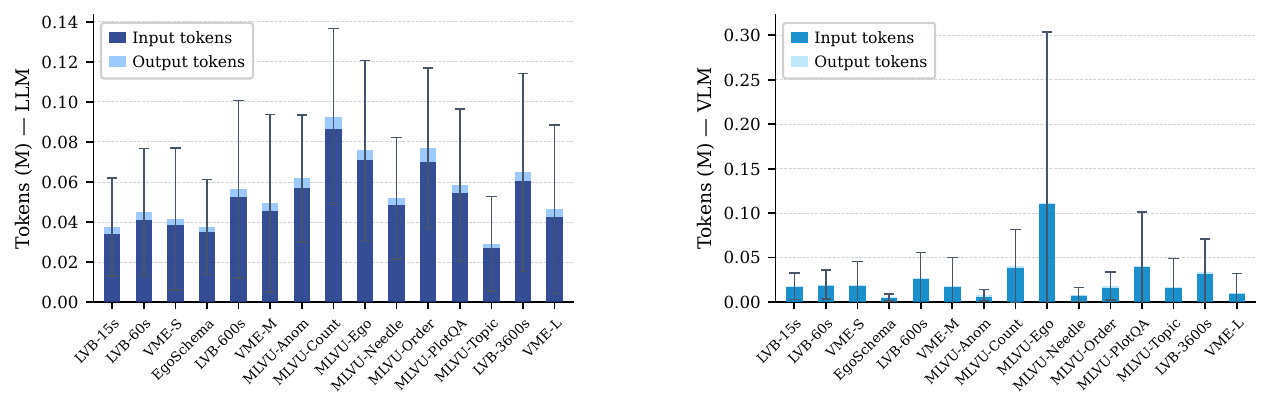}
    \caption{
    Breakdown of token usage for our method across benchmark dataset subsets. 
    The figure reports both input and output tokens for the main LLM and the VLM components, highlighting how our method distributes computation for different lengths and question types.
    }
    \label{fig:token_usage}
\end{figure*}

To better understand the computational cost of our method, we analyze the token usage associated with both the main LLM and the VLM components. Specifically, we report the input and output tokens consumed across different subsets of the benchmark datasets. The results are summarized in Figure~\ref{fig:token_usage}, which provides a detailed breakdown of token usage.

\subsection{Utilization Patterns of Probing Functions}

A key aspect of our framework is the set of probing functions that the main LLM invokes within the code execution environment to perform coding-based reasoning over videos. To better understand how the model utilizes these functions, we analyze the average usage frequency of probing functions across different question types and video lengths.

\begin{figure*}
    \centering
    \includegraphics[width=1\textwidth]{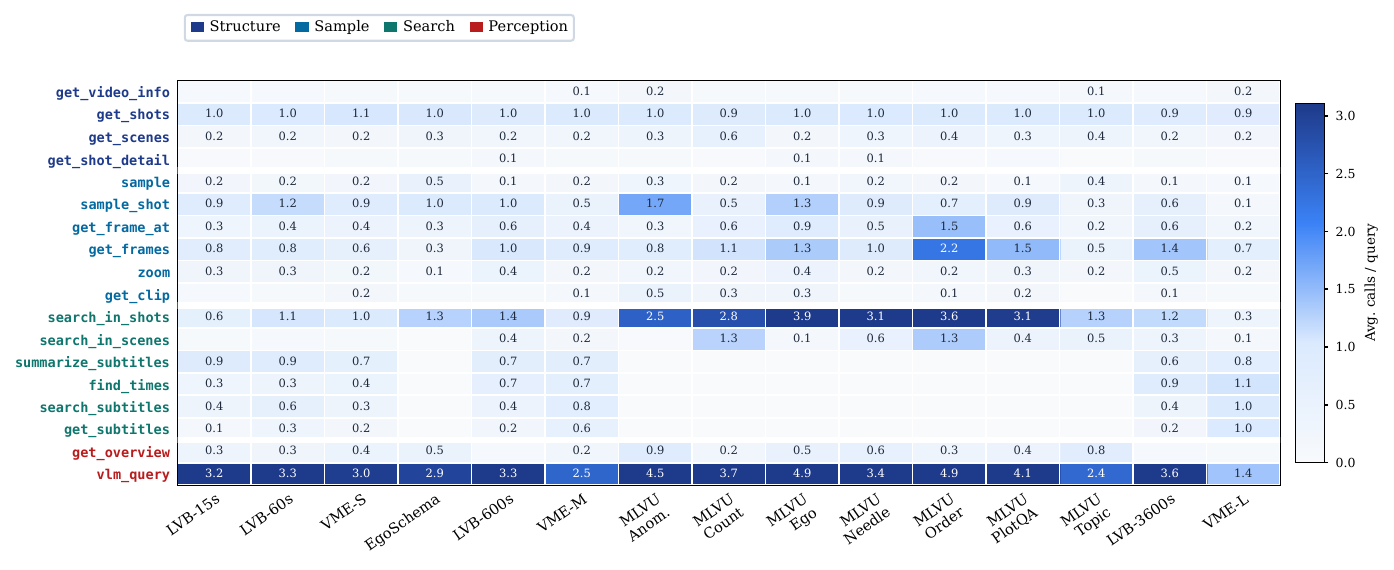}
    \caption{Average usage of probing functions across different question types and video lengths for four datasets used in our paper. Our framework predominantly relies on semantic search and VLM query functions, while other probing functions are invoked adaptively depending on the reasoning requirements of each question.}

    \label{fig:function_usage}
\end{figure*}

Figure~\ref{fig:function_usage} presents the distribution of function calls in our framework. The results show that the model frequently relies on \textit{semantic search} and \textit{VLM query} functions as its primary sources of information retrieval and visual reasoning. In addition, the usage of other probing functions varies depending on the question type, indicating that the model dynamically selects tools based on the reasoning requirements of each query.

\section{Extended Related Work}
\label{supp_sec:related_work}

\subsection{Exploiting Temporal Redundancy and Structured Video Representations}

Long videos contain substantial temporal redundancy, making direct frame-level processing inefficient. Video segmentation methods aim to partition videos into semantically or temporally coherent units such as shots or scenes, which serve as higher-level structural representations for downstream tasks~\cite{sun2025frames}. These segment-level abstractions enable more efficient indexing, retrieval, and reasoning compared to raw frame sequences.

A closely related area is video compression, which explicitly exploits temporal redundancy to reduce storage and transmission costs. Early compression techniques focused primarily on frame-level encoding, whereas modern video codecs incorporate motion estimation and motion compensation to model temporal dependencies across frames~\cite{musmann2005advances}. In typical compression pipelines, a subset of frames known as intra-coded frames (I-frames) are stored independently, while intermediate frames are reconstructed through motion vectors and residual information relative to reference frames. From an information-theoretic perspective, this process reduces redundancy by encoding only the incremental information necessary to reconstruct successive frames.
Although designed for compression, these representations implicitly capture important structural cues. For instance, motion vectors have been used as efficient approximations of optical flow, while I-frames often correlate with visually salient keyframes~\cite{ai2023key}. Prior work has leveraged such compression signals to accelerate video processing pipelines or approximate motion dynamics~\cite{wu2018compressed, jin2024video}.
In contrast to prior work that primarily uses compression artifacts for computational acceleration, we leverage temporal redundancy as a structural prior for hierarchical video decomposition. Our approach treats redundancy-aware segmentation as a first-stage abstraction layer that enables efficient downstream retrieval and reasoning over long videos.

\subsection{Moment Localization and Retrievalin Videos}

Video moment localization aims to identify temporal segments within long videos that correspond to a natural language query. Early approaches train task-specific architectures that align frame-level visual representations with textual embeddings through cross-modal objectives~\cite{anne2017localizing,Gao2017TALLTA,Wu2018MultimodalCF,Liu2018TemporalMN,Wang2020DualPI,liu2021survey}. These methods typically rely on densely sampled frames and specialized supervision signals designed for benchmark datasets.
More recent research has explored generalized retrieval frameworks that reduce reliance on task-specific training and improve cross-domain transferability~\cite{lin2023univtg,lei2021detecting,timechat2023}. These approaches aim to unify retrieval and localization tasks by learning more flexible video–text representations.

A closely related problem is text-based video retrieval, where the goal is to retrieve relevant videos given a natural language query. Early works extend contrastive image–text pretraining methods such as CLIP~\cite{radford2021learning} to the video domain by aggregating frame-level representations~\cite{zhang2024long0clip0,xu2021videoclip,wang2023internvid}. Subsequent models train large-scale video–text alignment systems specifically designed for video retrieval, achieving substantially stronger performance~\cite{wang2024videoclip,halbe2026virtue,wang2024internvideo20}.
Existing methods for language-based video localization often rely on dense captioning pipelines or classifier-based feature extraction~\cite{gao2017tall,huang2024vtimellm}. While effective for narrow tasks, these approaches typically require exhaustive frame-level encoding and predefined alignment procedures.

In contrast, our framework employs pretrained video–text retrieval models as semantic search modules within a hierarchical video representation. Rather than performing exhaustive frame-level matching, retrieval operates over structured temporal segments derived from our hierarchy, enabling scalable and query-adaptive localization within long videos.

\subsection{Long Video Understanding with Multimodal LLMs}

Understanding long-form videos remains a major challenge for multimodal large language models (MLLMs)~\cite{li2024llavaov,video-chat,mvbench2023,shen2024longvu}. Unlike short clips, long videos contain complex temporal dependencies, variable information density, and massive input sizes that strain current architectures.

One common strategy involves token compression techniques that reduce redundancy in visual representations~\cite{liless,hu2025thinking,yang2025visionzip,li2024llama,jiang2025storm}. These methods merge similar visual tokens or apply aggressive compression ratios to fit long videos within limited context windows~\cite{shen2024longvu,videollama}. However, such compression often discards fine-grained visual details, leading to degraded performance on tasks requiring precise temporal reasoning.

Another direction focuses on extending the context windows of large language models to support longer multimodal inputs~\cite{zhang2024longva}. While larger context windows enable models to process longer sequences, they do not inherently solve the challenge of efficiently locating relevant information within long videos.
Retrieval-based approaches attempt to address this limitation by selectively identifying informative segments. For example, VideoTree~\cite{wang2025videotree} constructs a hierarchical tree over frame-level representations to guide search through long videos. However, building such structures requires expensive frame-level clustering and captioning. Similarly, VCA~\cite{yang2025vca} formulates video exploration as a reinforcement learning problem where an agent navigates a graph representation of the video to identify relevant segments.

\begin{listing}
\begin{lstlisting}[style=mypython, numbers=none, caption={Code illustration
of Emergent Pattern 1. HPP adopts a coarse-to-fine exploration strategy using
inexpensive probing functions, integrating latent and code-based reasoning
with a verification loop to obtain the final answer. This mirrors
coarse-to-fine search strategies in classical computer
vision~\cite{burt1987laplacian} but is constructed entirely from probing
function compositions in the code.}, label={code:pattern1}]
# Question: What does the hand coming out of the computer do? A) Delivers a product B) Shakes the woman's hand C) Takes the woman's credit card D) Points at something on the screen.
# Stage 1 - coarse: search across shots with multiple phrasings
r1 = search_in_shots("hand coming out of computer", n=5)
r2 = search_in_shots("hand from screen reaching to woman", n=5)
results = (r1 or []) + (r2 or [])
seen = set()
unique = [r for r in results if not (seen.add(r['shot_id']) or r['shot_id'] in seen)]
# Stage 2 - After latent reasoning; fine: sample multiple frames from top-k candidate shots
frames = []
for sid in [37, 38, 39, 40]:
    frames += sample_shot(sid, n=2)
# Stage 3 - verify: single VLM call with answer options embedded
vlm_query("Which best describes the action: (A) Delivers a product, "
          "(B) Shakes the woman's hand, (C) Takes her credit card, (D) Points at something.",
          frames)
\end{lstlisting}
\end{listing}

Another line of work focuses on adaptive sampling and iterative refinement. The Thinking-with-Frames paradigm~\cite{maaz2024video,videochat2023} alternates between global reasoning and local inspection of candidate frames. SpecTemp~\cite{hu2025thinking} further decouples perception and reasoning through speculative temporal exploration, while other works employ hierarchical agent decomposition for targeted analysis~\cite{chen2025lvagent}.

\subsection{Agent-based Long Video Understanding.}
Recent approaches treat video understanding as an agentic process in which models interact with external tools to retrieve and analyze relevant information~\cite{wang2025videotree, yang2025vca, videoagent2024}. Early frameworks rely on retrieval tools such as CLIP~\cite{radford2021learning} to identify key frames~\cite{liless}, though they often struggle with domain gaps and limited temporal coverage. LVAgent~\cite{chen2025lvagent} introduces multi-agent collaboration among multimodal models, enabling dynamic team selection and iterative retrieval.
Our approach differs from these systems in two key aspects. First, we employ a single coding-capable LLM that performs strategy planning and iterative exploration within a programmatic execution REPL environment. Second, rather than relying on dense captioning pipelines or multiple specialized agents, our framework uses lightweight perceptual modules combined with hierarchical retrieval, enabling efficient and flexible reasoning over long videos.

\section{Emergent Coding and Probing Functions-based Strategy Patterns}
\label{supp_sec:emergent}

An important aspect of our method, HPP, is its ability to flexibly combine probing functions with a code execution environment (REPL) to analyze videos for efficient information extraction. This integration enables our method to iteratively query visual information and programmatically process intermediate results, leading to more structured and interpretable reasoning over complex video data.

In practice, we observe that HPP leverages these components in several unique and creative ways, forming distinct reasoning patterns during inference. To better understand these behaviors, we analyze the reasoning traces generated by our method and document representative samples here. 

To facilitate further inspection and reproducibility, we provide an interactive visualization tool in the attached code of supplementary material that allows readers to explore the reasoning traces in detail.

\subsection{Emergent Pattern 1: Hierarchical Coarse-to-Fine Localization}

As shown in Listing~\ref{code:pattern1}, our method is able to utilize hierarchical structure functions for coarse-to-fine localization. Rather than searching the full video at a uniform granularity, our method independently discovers a three-stage pyramid: (1) semantic search at the scene level to identify candidate temporal regions; (2) refinement at the shot level within those regions; (3) dense frame sampling within the top-scoring shots for VLM verification. This mirrors coarse-to-fine search strategies in classical computer vision~\cite{burt1987laplacian} but is constructed entirely from probing function compositions in the code. The VLM calls only on the most promising temporal regions. This emergent efficiency is never instructed. 

\subsection{Emergent Pattern 2: Multi-Query Ensemble Search with Deduplication}

For semantically ambiguous queries, our method (HPP) generates 5–15 paraphrase variants of the target concept (synonyms, compound descriptions, visual surface forms) and issues each as a separate \texttt{search\_in\_shots} call. Results are merged and deduplicated by \texttt{(start, end)} timestamp pair, forming a union-of-retrieval set that is substantially more complete than any single query alone. The explicit deduplication step demonstrates the utility of the code execution environment in the reasoning abilities of our method.

\begin{listing}
\begin{lstlisting}[style=mypython, numbers=none, caption={Illustration of
Pattern 2 -- Multi-Query Ensemble Search with Deduplication. Since semantic
search is nearly free after the first invocation (embeddings are cached), our
method (HPP) exploits this by issuing 5--15 paraphrase variants of the target
concept such as synonyms, compound descriptions, and visual surface forms, at
negligible extra cost. Then it utilizes REPL for results to be merged and
deduplicated by timestamp, forming a union-of-retrieval set more complete than
any single query alone.}, label={code:pattern2}]
# Question:  In this video, how many instances are there of the 'stomping grapes' action scene in total? A) 6 B) 0 C) 5 D) 1.
# Stage 2 - Localization
queries = ["grape stomping", "stomping", "grapes", "feet in grapes",
   "stepping on grapes", "wine press", "winery harvest"]
all_hits, seen = [], set()
for q in queries:
    for h in search_in_shots(q, n=30):
        key = (round(h['start'], 2), round(h['end'], 2))
        if key not in seen:
            seen.add(key)
            all_hits.append(h)
print(f"Total unique candidate shots: {len(all_hits)}")
\end{lstlisting}
\end{listing}

\subsection{Pattern 3: Subtitle-anchored Temporal Pinpointing}

When subtitles are available, our method (HPP) identifies that spoken text provides more precise temporal localization. It uses \texttt{find\_times()} to retrieve an exact timestamp for a target phrase, then uses that timestamp as a pointer for visual frame extraction (\texttt{zoom(t, context=2, n=8)}).

\begin{listing}
\begin{lstlisting}[style=mypython, numbers=none, caption={Code illustration of
Emergent Pattern 3 -- Subtitle-anchored Temporal Pinpointing. When subtitles
are available, HPP uses text as a precise temporal pointer: it searches for a
target phrase via regex to retrieve an exact timestamp, then uses that
timestamp to extract frames for VLM verification.}, label={code:pattern3}]
# Question: In the orange-red sunset, what is the girl, who is sitting back to the camera on a kayak with short sleeves and hair tied, doing when the subtitle 'exciting activities like kayaking and' appears? A) Admiring the sunset on the kayak B) Sitting on the kayak and waving her hands C) Chatting with friends D) Taking photos with a mobile phone E) Rowing the boat on the water.
# Stage 0 - subtitle pinpointing
matches = find_times(subtitles, r"exciting activities like kayaking and")
t = matches[0]['time']   # -> 106.13s
# Iteration 1 - visual extraction anchored to subtitle timestamp
frames = get_frames(t - 2, t + 2, n=6)
vlm_query("Is the girl facing away from camera on a kayak, "
          "wearing short sleeves with hair tied, under an orange-red sunset?", frames)
\end{lstlisting}
\end{listing}

\subsection{Pattern 4: Temporal Non-Maximum Suppression for Event Counting}

In counting tasks, our method (HPP) implements an explicit temporal clustering algorithm: (1) collect candidate event timestamps via multi-query search; (2) sort chronologically; (3) apply a gap-threshold merge (typically 1-2 s) to collapse adjacent shots that belong to the same event occurrence into a single cluster; (4) verify each cluster with VLM; (5) count verified clusters as the answer. This constitutes a spontaneous re-implementation of temporal non-maximum suppression. The 1-second gap threshold is chosen heuristically and varies across questions (0.5–2s depending on action type), suggesting it reasons about action granularity. This is a spontaneous re-implementation of temporal non-maximum suppression~\cite{brown2014temporal}.

\begin{figure*}
    \centering
    \includegraphics[width=1\linewidth]{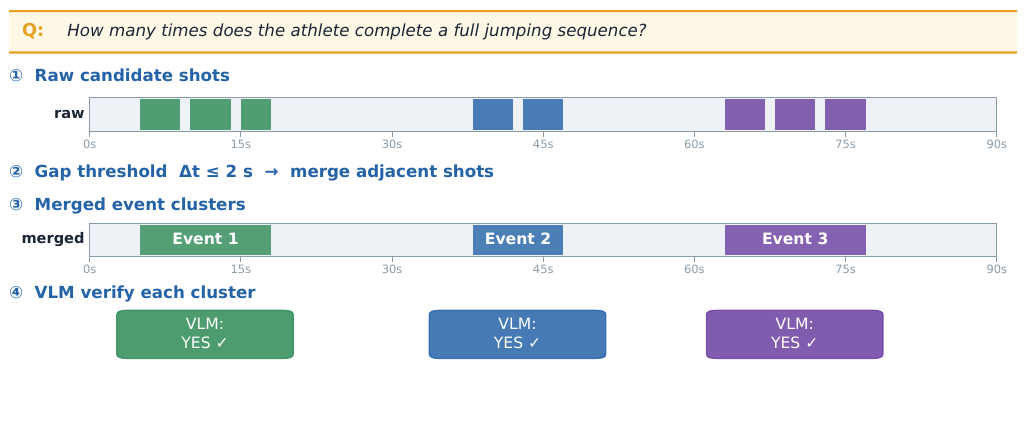}
    \caption{Code illustration of Emergent Pattern 4 - Temporal Non-Maximum Suppression for Event Counting. Our method (HPP) collects candidate event timestamps via multi-query semantic search, then applies a gap-threshold merge to collapse temporally adjacent shots into distinct event clusters using REPL. This is a spontaneous re-implementation of temporal non-maximum suppression~\cite{brown2014temporal}. Each cluster is then verified with VLM to yield a final count.}
    \label{fig:pattern4}
\end{figure*}

\subsection{Pattern 5: Structured Per-Shot VLM Verification with Output Parsing}

For both counting and ordering tasks, our method (HPP) designs a two-stage pipeline: semantic search returns O(20–50) candidate shots, then each shot is sent to the VLM (batched) with a binary or structured-output prompt ("YES - playing trombone (count=N) | NO"). It then parses VLM natural language responses using \texttt{re.search()} to extract structured fields, enabling programmatic aggregation of VLM extracted information across many calls. This constitutes a spontaneous form of tool-use augmentation: treating the VLM as a structured classifier rather than a free-form reasoner, and using \texttt{re.search} to recover structured data from its outputs. An example is shown in Code Listing~\ref{lst:pattern5}.

\begin{listing}
\begin{lstlisting}[style=mypython, numbers=none, caption={Code illustration of
Emergent Pattern 5. Structured Per-Shot VLM Verification with Output Parsing.
HPP constrains the VLM to a fixed response format and parses it with a regex,
turning free-form visual judgments into structured, countable results that can
be aggregated programmatically across shots.}, label={lst:pattern5}]
scene_yes = []
for sid in candidate_scenes:
    sc  = next(s for s in scenes if s['scene_id'] == sid)
    imgs = zoom(sc, context=0.1, n=6)
    resp = vlm_query(
        f"Scene {sid} [{sc['start']:.1f}-{sc['end']:.1f}s]: "
        "Is any person actively playing a trombone? "
        "Answer ONLY: 'Yes -- playing trombone (count=N)' or 'No'.", imgs)
    if resp.lower().startswith("yes"):
        m = re.search(r'count\s*=\s*(\d+)', resp)
        scene_yes.append({'scene_id': sid, 'count': int(m.group(1)) if m else 1})
\end{lstlisting}
\end{listing}

\subsection{Pattern 6: Parallel Event Localization for Temporal Ordering}

For chronological ordering questions, our method (HPP) decomposes the global ordering problem into N independent localization subproblems: one per event in the ordering question. Each event is located via a dedicated semantic search call. The representative timestamp of each event's top search hit is then used for comparison, converting temporal ordering into simple integer comparison. An example code listing is shown in Listing~\ref{lst:pattern6}.

\begin{listing}
\begin{lstlisting}[style=mypython, numbers=none, caption={Code illustration of
Emergent Pattern 6. Parallel Event Localization for Temporal Ordering. Our
method decomposes the global ordering problem into N independent localization
subproblems and uses semantic search for localization.}, label={lst:pattern6}]
# One search call per event: parallel localization
r_abseil  = search_in_shots("abseiling, rappelling down cliff", n=5)   # -> 2.0s
r_jewelry = search_in_shots("making jewelry, crafting necklace", n=5)  # -> 132.9s
r_milking = search_in_shots("milking cow, dairy", n=5)                 # -> 195.4s
r_toilet  = search_in_shots("cleaning toilet, bathroom", n=5)          # -> 292.2s
# Verify each with a single keyframe, then order by timestamp
order = sorted([('abseil',2.0), ('jewelry',132.9),
                ('milking',195.4), ('toilet',292.2)], key=lambda x: x[1])
# -> abseil -> jewelry -> milking -> toilet  ->  Answer: A
\end{lstlisting}
\end{listing}

\section{Additional Implementation Details}

\noindent \textbf{General Details. }For our experiments, we use GPT-5-mini as the main LLM, Qwen-3V-L 2B~\cite{qwenvl-2023} as the VLM, and VideoXL~\cite{videoxl} for semantic video search. Unless explicitly stated otherwise, results for prior methods are extracted directly from the corresponding publications. To ensure a fair comparison, we run the baseline VLM at the same frame sampling rate used in our method, aligning input conditions across all evaluations.

\noindent\textbf{System Prompt. }
\label{supp_sec:sys_prompt}
In this section, we briefly discuss the system prompt used in our framework. As discussed in the main paper, the prompt is built upon the Recursive Language Model (RLM) framework~\cite{zhang2025recursive} and is adapted for long-form video understanding. Specifically, we extend the original formulation with video-specific instructions and structured guidance that help the model plan exploration strategies and efficiently interact with the video through the REPL.

The prompt begins with a high-level description of the task and the overall reasoning setting. It then introduces the REPL environment, including the accessible variables and the available probing functions that the model can invoke to inspect different parts of the video. In addition, the prompt provides several strategy hints and operational guidelines that encourage the model to reason about the video structure, progressively refine its search, and avoid unnecessary visual queries. Finally, the prompt includes a set of explicit rules together with a short example illustrating how the probing functions can be combined during reasoning. These components guide the LLM to formulate structured exploration strategies and effectively navigate long videos during inference.

\end{document}